\documentclass{article}

\usepackage[preprint]{neurips_2025}

\usepackage[utf8]{inputenc} % allow utf-8 input
\usepackage[T1]{fontenc}    % use 8-bit T1 fonts
\usepackage{hyperref}       % hyperlinks
\usepackage{url}            % simple URL typesetting
\usepackage{booktabs}       % professional-quality tables
\usepackage{nicefrac}       % compact symbols for 1/2, etc.
\usepackage{pifont}         % for \ding symbols
\usepackage{microtype}      % microtypography
\usepackage[table,xcdraw]{xcolor}        % colors
\usepackage{bbding}
\usepackage{natbib}

\usepackage{amsmath,amssymb,amsfonts}
\usepackage{algorithmic}
\usepackage{graphicx}
\usepackage{textcomp}
\usepackage{array}
\usepackage{utfsym}

\usepackage{multirow} % tables
\usepackage{colortbl}
\usepackage{enumitem}
\usepackage{makecell}
\usepackage{enumitem}
\usepackage{amsmath,amsfonts,bm}

\def\eqref#1{equation~\ref{#1}}
\def\1{\bm{1}}

\DeclareMathAlphabet{\mathsfit}{\encodingdefault}{\sfdefault}{m}{sl}
\SetMathAlphabet{\mathsfit}{bold}{\encodingdefault}{\sfdefault}{bx}{n}

\newcommand{\eg}{\textit{e}.\textit{g}.}
\newcommand{\ie}{\textit{i}.\textit{e}.}

\newcounter{bxincomm}
\definecolor{aqua}{rgb}{0.00,0.67,0.80}
\usepackage{colortbl}
\definecolor{gold1}{RGB}{100,200,100}  % Top 1
\definecolor{gold2}{RGB}{140,220,140}  % Top 2
\definecolor{gold3}{RGB}{180,240,180}  % Top 3
\definecolor{gold4}{RGB}{220,255,220}  % Top 4

\newcounter{todocomm}

\newcommand{\venus}{\textsc{VenusX}}

\title{\venus: Unlocking Fine-Grained Functional Understanding of Proteins}

\author{%
Yang Tan$^{1,2}$\thanks{Equal Contribution. Corresponding to: Bingxin Zhou (\texttt{bingxin.zhou@sjtu.edu.cn})} 
\quad Wenrui Gou$^{2*}$ 
\quad Bozitao Zhong$^{1}$ 
\quad Liang Hong$^{1}$ 
\quad Huiqun Yu$^{2}$ \\
\quad \textbf{Bingxin Zhou}$^{1}$ \\
$^1$ Shanghai Jiao Tong University  \\ 
$^2$ East China University of Science and Technology \\
}

\begin{document}

\maketitle

\begin{abstract}
Deep learning models have driven significant progress in predicting protein function and interactions at the protein level. While these advancements have been invaluable for many biological applications such as enzyme engineering and function annotation, a more detailed perspective is essential for understanding protein functional mechanisms and evaluating the biological knowledge captured by models. To address this demand, we introduce \venus, the first large-scale benchmark for fine-grained functional annotation and function-based protein pairing at the residue, fragment, and domain levels. \venus~comprises three major task categories across six types of annotations, including residue-level binary classification, fragment-level multi-class classification, and  pairwise functional similarity scoring for identifying critical active sites, binding sites, conserved sites, motifs, domains, and epitopes. The benchmark features over $878,000$ samples curated from major open-source databases such as InterPro, BioLiP, and SAbDab. By providing mixed-family and cross-family splits at three sequence identity thresholds, our benchmark enables a comprehensive assessment of model performance on both in-distribution and out-of-distribution scenarios. For baseline evaluation, we assess a diverse set of popular and open-source models, including pre-trained protein language models, sequence-structure hybrids, structure-based methods, and alignment-based techniques. Their performance is reported across all benchmark datasets and evaluation settings using multiple metrics, offering a thorough comparison and a strong foundation for future research. Code and data are publicly available at \url{https://github.com/ai4protein/VenusX}.

\end{abstract}

\section{Introduction}
Deep learning has significantly advanced the analysis of large-scale protein data, enabling efficient solutions to key inference tasks across sequence, structure, and function. Notable successes include structure prediction \cite{jumper2021alphafold2,abramson2024alphafold3}, sequence engineering \cite{lu2022machine,zhou2024cpdiffusion,tan2024venusrem}, and functional annotation \cite{yu2023clean,zhou2024fsfp}. The rapid progress in this field is supported not only by the models’ scientific and practical value, but also by the availability of high-quality benchmarks that define clear learning objectives and ensure fair, reproducible evaluation.

A wide range of datasets and evaluation protocols have been developed to facilitate model training and assessment, especially those centered on large-scale protein sequence and structure data \citep{cath1997,varadi2022alphafolddb,uniprot2025uniprot}. While some benchmarks include functional annotations, they predominantly target protein-level properties, where the goal is to assign a single label to an entire protein or protein pair. Some representative tasks include function annotation \citep{tan2024sesadapter,li2025VenusVaccine}, protein–protein interaction prediction \citep{pan2010human_ppi,szklarczyk2019string,jankauskaite2019skempi2,moine2024ppi_dataset}, and protein fitness estimation \citep{gray2018envision,riesselman2018deepsequence,notin2024proteingym,zhang2025venusmuthub}.

Despite the overwhelming focus on protein-level benchmarks, biological functions are often governed by specific subregions within proteins rather than the entire molecule. Global labels can obscure mechanistic details and may even lead models to rely on biologically implausible features for prediction. This increases the risk of overfitting to noise, reduces interpretability, and compromises accuracy in tasks where local features are critical, such as function annotation \citep{cagiada2023discovering, lee2007predicting} and paratope design \citep{attique2023deepbce}. As a result, \textbf{there is a growing demand for benchmarks that support supervision and evaluation at a fine-grained resolution}. Such resources are essential not only for advancing functional understanding but also for systematically assessing how well learned representations capture biologically meaningful signals beyond sequence similarity.

We address this gap by introducing \venus, the first large-scale and biologically grounded benchmark for fine-grained protein understanding. \venus~spans multiple subprotein levels—including residues, motifs, fragments, and domains—and is designed to evaluate model performance across three task categories: (1) \textbf{residue-level binary classification}, which assesses whether individual amino acids contribute critically to protein function, such as catalysis, ligand binding, evolutionary constraint, or domain boundaries; (2) \textbf{fragment-level multi-class classification}, which identifies functional subregions within a protein and assigns them to specific biological roles; and (3) \textbf{pairwise functional similarity scoring}, which matches functionally similar proteins or substructures without requiring explicit function labels.

The raw residue-level annotations are sourced from three high-quality databases: \textit{InterPro} \citep{paysan2023interpro}, \textit{BioLiP} \citep{yang2012biolip}, and \textit{SAbDab} \citep{dunbar2014sabdab}. We curate over $878,000$ high-confidence samples, which form the basis of diverse tasks across three categories of fine-grained functional prediction. To enable comprehensive evaluation of model fitness, robustness, and generalizability, we consider both label distribution and input similarity at the fragment and protein levels and define multiple evaluation setups with different partitioning strategies for training and testing.

We benchmark a broad spectrum of popular protein representative models to assess their effectiveness on \venus. These include pre-trained protein language models \citep{rives2021esm1b, elnaggar2021prottrans, elnaggar2023ankh, heinzinger2023prostt5,lin2023esm2, hamamsy2024tmvec}, sequence-structure hybrid models \citep{su2023saprot, tan2025protssn}, inverse folding models \citep{mifst, hsu2022esm-if1}, structure-based geometric networks \citep{gvp}, and traditional alignment-based methods \citep{altschul1990blast, zhang2005tmalign, van2024foldseek}. We observe substantial variation in performance across annotation types, sequence identity thresholds, and task formulations. These findings reveal that strong performance on conventional global protein-level tasks does not necessarily translate to fine-grained functional understanding. The results further suggest that many current models rely heavily on global or distributional cues, rather than capturing precise, localized biological signals. These limitations highlight the need for future model designs that are better aligned with the demands of fine-grained benchmarks, which emphasize robustness, generalization across protein families, and biological interpretability.

In summary, \venus~establishes the first large-scale benchmark for fine-grained protein understanding, featuring diverse tasks built on curated residue-level annotations. It offers biologically meaningful evaluation dimensions for future protein models and enables systematic assessment of their ability to capture true biological knowledge. All datasets, task definitions, evaluation protocols, and baseline leaderboards are publicly available to support followup research in the future.

\begin{figure*}[t!]
    \centering
    \includegraphics[width=\textwidth]{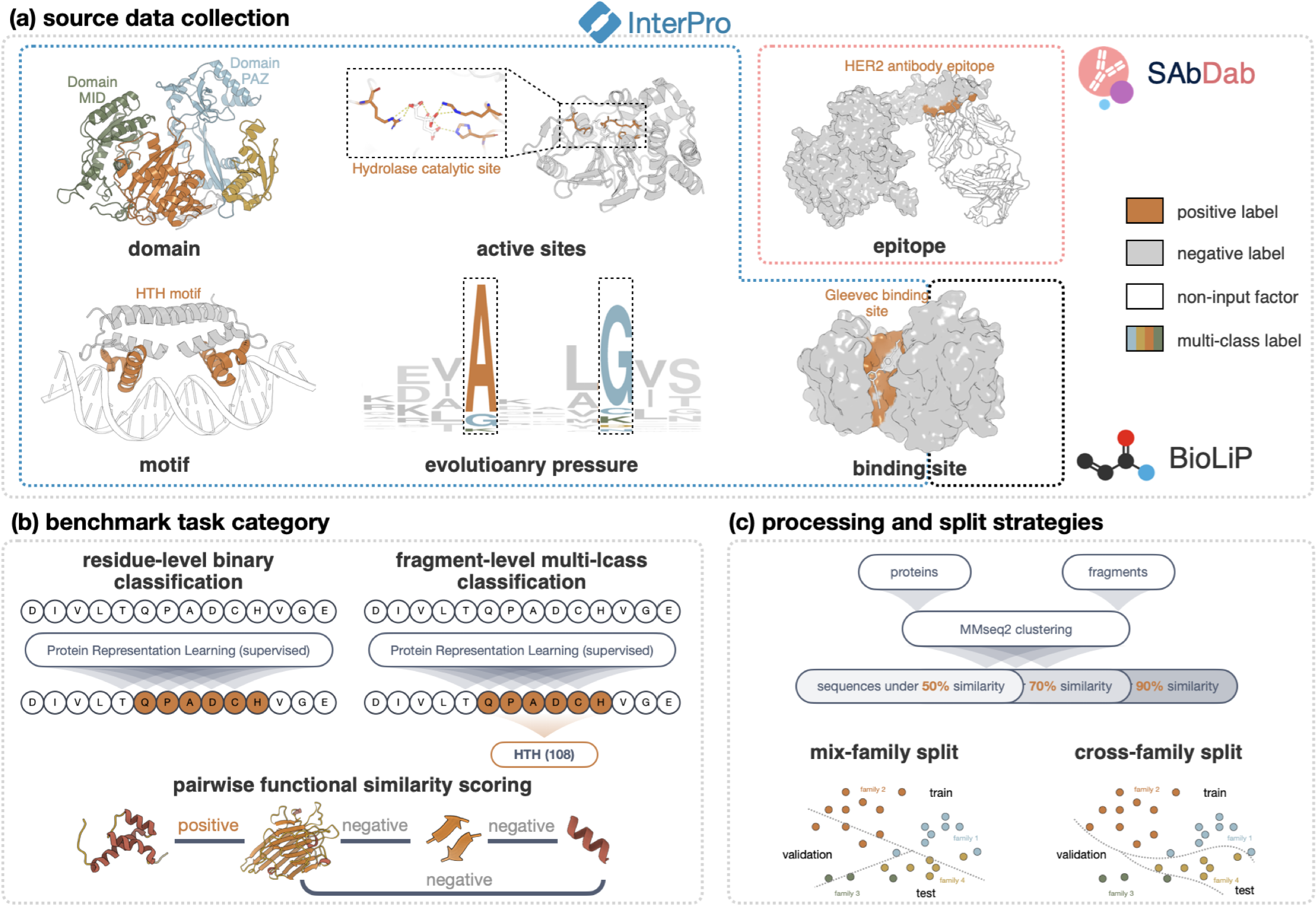}
    \vspace{-3mm}
    \caption{Overview of the \venus~benchmark. (a) Six types of functional annotations collected from InterPro, BioLiP, and SAbDab (Section~\ref{sec:source}). (b) Three benchmark task categories: residue-level and fragment-level classification, and pairwise similarity scoring (Sections~\ref{sec:cat1Task}-\ref{sec:cat3Task}). (c) Sequence identity–based clustering and mix-family and cross-family data split strategies (Sections~\ref{sec:partition}).}
    \label{fig:framework}
\end{figure*}

\section{Data Collection and Curation}
\label{sec:source}
We collect residue- and fragment-level annotations from InterPro \citep{paysan2023interpro}, BioLiP \citep{yang2012biolip}, and SAbDab \citep{dunbar2014sabdab}, followed by thorough cleaning, redundancy removal, identity-based clustering, and alignment with structure and annotations.

\subsection{InterPro: Functional Annotations Across Diverse Protein Families}
We use InterPro \citep{paysan2023interpro} to collect residue-level entries from various annotation categories, such as active sites, binding sites, and functional domains. For each category, metadata are retrieved from \url{https://www.ebi.ac.uk/interpro/} in \texttt{.json} format, including InterPro family identifiers, associated Gene Ontology (GO) terms, and annotated residue positions. UniProt identifiers and their corresponding functional annotations are extracted using VenusFactory \citep{tan2025venusfactory}. Canonical protein sequences are obtained from UniProt \citep{uniprot2025uniprot}, and predicted structures are retrieved from the AlphaFold Protein Structure Database \citep{varadi2022alphafolddb}, retaining only entries with available structural models.

To ensure non-redundancy, we remove additional entries with identical annotated fragments. If a protein contains multiple distinct fragments annotated with the same function, the annotations are consolidated into a single entry. The resulting dataset comprises curated, functionally annotated fragments with aligned sequence–structure pairs, suitable for fine-grained residue-level analysis.

\subsection{BioLiP: Experimentally-Derived Ligand Binding Sites}
We incorporate additional residue-level binding site annotations from BioLiP \citep{yang2012biolip}, a curated resource of protein–ligand interactions derived from experimentally resolved complexes in the Protein Data Bank (PDB) \citep{burley2019rcsb}. Binding residues are identified using a distance-based criterion. A residue is labeled as part of the binding site if any of its atoms lie within the sum of the Van der Waals radii of the interacting atom pair plus a $0.5$ \AA~empirical margin. This approach captures steric interactions and enables robust identification of physiologically relevant binding interfaces.

For each entry, we extracted the complete receptor sequence, its PDB chain identifier, the annotated binding site residues, and the ligand identity specified by its Chemical Component Dictionary (CCD) code. Sequence and coordinate information were parsed directly from the corresponding PDB files, and ligand annotations were stringently curated to retain only biologically relevant molecules. This curated dataset augments the InterPro‑derived entries with high‑resolution protein–ligand binding‑site pairs, providing a biologically meaningful benchmark for training and evaluating deep learning models on protein-ligand complexes.

\subsection{SAbDab: Structure-Based Epitope Annotations for Antibody–Antigen Complexes}
We extract epitope-level annotations from antibody–antigen complexes in SAbDab \citep{dunbar2014sabdab}, with a restriction to entries where the antigen is a protein. Metadata are sourced from curated \texttt{.tsv} files from \url{https://opig.stats.ox.ac.uk/webapps/sabdab-sabpred/sabdab/search/}, including antibody heavy and light chains, as well as the associated antigen chains. Structures are parsed using BioPython \citep{cock2009biopython}. We only retain standard residues with defined $C_\alpha$ coordinates. An antigen residue is considered part of the epitope if the Euclidean distance between its $C_\alpha$ atom and any antibody $C_\alpha$ atom is less than $10$ \AA. This geometric criterion captures both continuous epitopes (adjacent in sequence) and conformational epitopes (spatially clustered but sequence-distant).

We treat antigens with identified epitope residues from this procedure as valid entries. For each of them, we extract the complete amino acid sequence, the epitope residue indices (start from 0), and associated structure files. Entries with non-protein antigens, missing chain information, or structural inconsistencies are excluded. The resulting dataset offers structure-derived epitope labels aligned with sequence data to support targeted development and evaluation of immune-related protein models.

\begin{table*}[t]
\caption{Summary of the 7 residue-level classification tasks. For sequence length, number of positive residues, and proportion of positives, averages are shown with standard deviations in parentheses.}
\label{tab:residue-task}
\centering
\resizebox{\textwidth}{!}{%
    \begin{tabular}{llrrrrcc}
    \toprule
    \textbf{Target} & 
    \textbf{Description} & \textbf{\# Proteins} & \textbf{Seq. Len.} & \textbf{\# Positive} & \textbf{\% Positive} & \textbf{Cross} & \textbf{Mix} \\ 
    \midrule
    \textbf{Act} & active sites (InterPro) & $9,667$ & $482.5\;_{(349.0)}$ & $16.7\;_{(7.0)}$ & $4.6\;_{(2.8)}$ & \Checkmark & \Checkmark \\
    \textbf{BindI} & binding sites (InterPro) & $8,959$ & $486.5\;_{(339.8)}$ & $24.5\;_{(20.9)}$ & $7.9\;_{(7.3)}$ & \Checkmark & \Checkmark \\
    \textbf{BindB} & binding sites (BioLiP) & $115,505$ & $348.7\;_{(272.9)}$ & $9.4\;_{(9.6)}$ & $4.1\;_{(5.7)}$ & & \Checkmark \\
    \textbf{Evo} & evolutionary pressure (InterPro) & $59,948$ & $365.9\;_{(290.8)}$ & $23.4\;_{(13.1)}$ & $10.4\;_{(9.2)}$ & \Checkmark & \Checkmark \\
    \textbf{Motif} & motif (InterPro) & $10,271$ & $595.2\;_{(383.5)}$ & $78.0\;_{(73.4)}$ & $20.2\;_{(22.8)}$ & \Checkmark & \Checkmark \\
    \textbf{Dom} & domain (InterPro) & $595,454$ & $537.5\;_{(373.2)}$ & $169.2\;_{(117.2)}$ & $40.3\;_{(26.1)}$ & \Checkmark & \Checkmark \\
    \textbf{Epi} & epitope (SAbDab) & $5,370$ & $374.9\;_{(291.3)}$ & $24.8\;_{(10.5)}$ & $10.6\;_{(9.2)}$ & & \Checkmark \\ \bottomrule
    \end{tabular}%
}

\end{table*}

\section{Benchmark Tasks}
\subsection{Residue-Level Binary Classification}
\label{sec:cat1Task}
\paragraph{Task Description}
The first category of tasks focuses on identifying functionally important residues within a protein. Each task is framed as a binary classification problem at the residue level, where positions are labeled as either functionally relevant (positive) or irrelevant (negative). Related functions involve catalysis, binding, or other biological processes, based on curated annotations from the sources in Section~\ref{sec:source}. Table~\ref{tab:residue-task} summarizes all $7$ subtasks. Unlike conventional protein-level function prediction, these tasks assess whether models can detect critical residues independent of global function or family context. They encourage models to go beyond coarse representations and capture residue-level signals that may support model analysis in terms of interpretability or explanability. In practice, accurate residue-level predictions can aid enzyme engineering, mutational effect analysis, and active-site redesign.

\paragraph{Evaluation Metrics}
These tasks exhibit strong class imbalance, as most residues are non-functional (see ``\% Positive" in Table~\ref{tab:residue-task}). To better assess model performance on the minority class, we prioritize metrics that emphasize positive predictions. Specifically, we report class-specific precision, recall, and F1 score for the positive class, along with the area under the precision–recall curve (AUPR), which offers a more informative evaluation than ROC-AUC under imbalanced conditions.

\subsection{Fragment-Level Multi-Class Classification}
\paragraph{Task Description}
The second category of tasks moves from residue-level identification to fragment-level classification. Each of $5$ tasks is defined as a multi-class classification problem that involves assigning contiguous, annotated regions to their corresponding InterPro families (see Table~\ref{tab:fragment-task}). This reflects a practical two-stage inference pipeline that first locates functionally relevant regions, then assigns them to functional families. Since proteins often contain multiple functional fragments, the task also supports multi-label (InterPro family) annotations and encourages models to capture compositional functionality. It also bridges residue-level predictions with interpretable biological labels grounded in existing ontologies. This task category serves as a useful testbed for evaluating whether learned representations support compositional reasoning over functional subunits within proteins. This is particularly relevant for modular or multifunctional proteins, enabling applications such as domain annotation, structural proteomics study, and drug target discovery.

\paragraph{Evaluation Metrics}
These tasks assign each functional fragment to its corresponding InterPro family, which may include hundreds to tens of thousands of distinct classes (see ``\# Class” in Table~\ref{tab:fragment-task}). We report accuracy (ACC), macro-averaged precision, recall, F1 score, and Matthews correlation coefficient (MCC). We take ACC and macro-F1 as the primary metrics to reflect both overall correctness and class-balanced performance.

\begin{table}[t]
\centering
\begin{minipage}{0.46\textwidth}
\centering
\caption{Summary of the 5 fragment-level multi-class classification tasks. }
\label{tab:fragment-task}
\resizebox{\textwidth}{!}{%
    \begin{tabular}{lrrr}
    \toprule
    \textbf{Target} & \textbf{\# Fragments} & \textbf{Seq. Len.} & \textbf{\# Class} \\
    \midrule
    \textbf{Act} & $9,767$ & $18.7\;_{(7.0)}$ & $132$ \\
    \textbf{BindI} & $10,562$ & $26.6\;_{(21.7)}$ & $76$ \\
    \textbf{Evo} & $66,916$ & $25.5\;_{(13.3)}$ & $740$ \\
    \textbf{Motif} & $13,244$ & $80.23\;_{(73.8)}$ & $454$ \\
    \textbf{Dom} & $656,669$ & $171.3\;_{(117.4)}$ & $13,459$ \\
    \bottomrule
    \end{tabular}
}
\end{minipage}
\hfill
\begin{minipage}{0.51\textwidth}
\centering
\caption{Summary of the five pairwise similarity scoring tasks (in millions). %“\# Full-P/N” and “\# Frag-P/N” denote the total number of positive and negative pairs sampled by full sequences or fragments within InterPro families.
}
\vspace{-2mm}
\label{tab:unsup-task}
\resizebox{\textwidth}{!}{%
    \begin{tabular}{lrrrr}
    \toprule
    & \multicolumn{2}{c}{\textbf{Protein}} & \multicolumn{2}{c}{\textbf{Fragment}} \\ \cmidrule(lr){2-3}\cmidrule(lr){4-5}
    \textbf{Target} & \textbf{\# Positive} & \textbf{\# Negative} & \textbf{\# Positive} & \textbf{\# Negative} \\
    \midrule
    \textbf{Act} & $1.3$ & $45.4$ & $1.3$ & $46.4$ \\
    \textbf{BindI} & $3.5$ & $36.6$ & $5.0$ & $50.8$ \\
    \textbf{Evo} & $7.7$ & $1,789.1$ & $10.0$ & $2,228.9$ \\
    \textbf{Motif} & $2.4$ & $50.3$ & $6.0$ & $81.7$ \\
    \textbf{Dom} & $217.7$ & $177,064.8$ & $346.0$ & $215,260.7$  \\
    \bottomrule
    \end{tabular}
    }
\end{minipage}
\end{table}

\subsection{Pairwise Functional Similarity Scoring}
\label{sec:cat3Task}
\paragraph{Task Description}
The third task category evaluates how well models capture meaningful similarities between proteins or fragments without supervision. Given a pair of inputs (either proteins or fragments), the goal is to assess their functional relatedness based on their similarity in sequence, structure, or embedding. Ground-truth labels indicate whether the pair belongs to the same InterPro family. This task provides a retrieval-style evaluation of representation quality, particularly for identifying subtle but biologically relevant similarities. It has important practical value in applications such as enzyme mining, remote homolog detection, and functional clustering in metagenomic datasets. A full dataset summary is provided in Table~\ref{tab:unsup-task}.

\paragraph{Evaluation Metrics}
Pairwise similarity scoring tasks are evaluated using the area under the ROC curve (AUC). We use cosine similarity between protein representations as the similarity score for embedding-based methods. For alignment-based methods that explicitly compute sequence or structure alignments, we employ the negative logarithm of the E-value (\eg, for \textsc{Foldseek} \citep{van2024foldseek} and \textsc{BLAST} \citep{altschul1990blast}) or the bi-directional average TM-score (\eg, for \textsc{TM-align} \citep{zhang2005tmalign}).

\subsection{Partitioning Protocol for Training and Evaluation}
\label{sec:partition}
\paragraph{Classification Tasks}
We evaluate both in-distribution and out-of-distribution prediction performance. To this end, we construct mix-family and cross-family data splits for both residue-level and fragment-level tasks.
(1) Mix-family splits assess in-distribution generalization by randomly partitioning proteins (or fragments) into training, validation, and test sets in an 8:1:1 ratio, without considering family assignments.
(2) Cross-family splits evaluate out-of-distribution generalization by assigning entire InterPro families to training, validation, and test sets in the same 8:1:1 ratio.
For both strategies, we first apply MMseqs2 clustering \citep{steinegger2017mmseqs2} at 50\%, 70\%, and 90\% sequence identity thresholds to reduce redundancy before splitting. Note that, due to limitations in available data and family annotations across the original source databases, mix-family splits on proteins are applied to datasets from all three sources. In contrast, fragment-level splits and cross-family splits based on family identity are applied only to InterPro-sourced datasets. Detailed availability and train/validation/test set statistics are provided in Tables~\ref{app:tab:residue-split-50}–\ref{app:tab:residue-split-70-90} in Appendix~\ref{app:split_detail}.

\paragraph{Similarity Scoring Tasks}
As the third task category of pairwise functional similarity scoring does not involve model supervision, we do not perform data partitioning. Instead, following a similar strategy to \cite{tan2024protloca}, we uniformly subsample a set of positive and negative pairs from the complete dataset for evaluation. This random subsampling is necessary due to the combinatorially large number of possible protein pairs in the similarity scoring task (for instance, see Table~\ref{tab:unsup-task}). Specifically, for both pairing tasks on fragments and proteins, we randomly sample $10,000$ positive pairs (\ie, proteins from the same InterPro family) and $10,000$ negative pairs (\ie, proteins from different InterPro families) as one evaluation dataset. We repeat the procedure using three different random seeds, and the final performance scores are averaged across the three repetitions.

\subsection{Naming Protocol for Benchmark Datasets}
Following the construction options introduced in Sections~\ref{sec:cat1Task}-\ref{sec:partition}, \venus~includes a total of 56 datasets.  For clarity, each dataset is named by \texttt{\venus\_[category]\_[target]\_[split]}, where each part denotes the task category, prediction target, and data split strategy.
\begin{itemize}[leftmargin=*]
    \item \texttt{[category]} refers to the task category, with three choices of \texttt{Res}, \texttt{Frag}, and \texttt{Pair} to represent Residue-level tasks, fragment-level tasks, and pairwise scoring tasks.
    \item \texttt{[target]} represents the $7$ cases of targets, including \texttt{Act} (active sites), \texttt{BindI} (binding sites from InterPro), \texttt{BindB} (binding sites from BioLiP), \texttt{Evo} (evolutionary pressure), \texttt{Motif} (functional motif), \texttt{Dom} (functional domain), and \texttt{Epi} (epitope sites).
    \item \texttt{[split]} denotes the partitioning strategies, with \texttt{X} for cross-family and \texttt{M} for mix-family splits, \texttt{P} and \texttt{F} indicating protein or fragment, and the final number representing the clustering threshold.
\end{itemize}
For instance: (1) \texttt{\venus\_Res\_BindB\_X} is a residue-level binary classification task that predicts binding sites (from BioLiP), using a cross-family split. (2) \texttt{\venus\_Frag\_Act\_MF90} denotes a fragment-level multi-class classification task targeting active sites, with a mix-family split on fragment entries clustered at $90\%$ identity.

\section{Experiments}\label{sec:exp}

\begin{table}[t]
\caption{
Residue-level classification performance across datasets and data splits. “MF50” and “MP50” refer to mixed-family splits with 50\% sequence identity filtering applied to fragments and proteins, respectively. \textbf{Top-1}, \underline{Top-2}, and \textit{Top-3} results for each target dataset are highlighted, respectively. Models are grouped by input modality. AUPR scores for each task are reported, and detailed results are provided in Tables~\ref{app:tab:token_cls_detail_results_1}-\ref{app:tab:token_cls_detail_results_3} of the Appendix~\ref{app:add_exp_sec}.}
\label{tab:residue-cls}
\centering
\resizebox{\textwidth}{!}{%
    \begin{tabular}{@{}llcccccccc@{}}
    \toprule
    \multirow{2}{*}{\textbf{Target}} & \multirow{2}{*}{\textbf{Split}} 
    & \multicolumn{4}{c}{\textbf{Sequence-only}} 
    & \multicolumn{3}{c}{\textbf{Sequence-Structure}} 
    & \multicolumn{1}{c}{\textbf{Structure-only}} \\
    \cmidrule(lr){3-6} \cmidrule(lr){7-9} \cmidrule(lr){10-10}
     & 
     & \textsc{ESM2-t30} & \textsc{ESM2-t33} & \textsc{ProtBert} & \textsc{Ankh-base} 
     & \textsc{SaProt-35M} & \textsc{SaProt-650M} & \textsc{ProtSSN} 
     & \textsc{GVP-GNN} \\
    \midrule
    \multirow{3}{*}{\textbf{Act}}
    & MF50  & \underline{0.855} & \textit{0.852} & 0.764 & \textbf{0.873} & 0.688 & 0.745 & 0.465 & 0.523 \\
    & MP50  & 0.932 & \underline{0.955} & 0.895 & \textbf{0.960} & 0.905 & \textit{0.945} & 0.917 & 0.898 \\
    & Cross       & 0.143 & 0.143 & 0.131 & \underline{0.166} & 0.114 & \textbf{0.185} & \textit{0.156} & 0.101 \\
    \midrule
    \multirow{3}{*}{\textbf{BindI}}
    & MF50  & \textbf{0.912} & \textit{0.904} & 0.857 & \underline{0.907} & 0.807 & 0.838 & 0.801 & 0.611 \\
    & MP50  & \textit{0.963} & \textbf{0.971} & 0.926 & \underline{0.970} & 0.927 & 0.960 & 0.907 & 0.883 \\
    & Cross       & 0.133 & \textit{0.159} & 0.112 & 0.145 & \textbf{0.230} & \underline{0.182} & \underline{0.182} & 0.040 \\
    \midrule
    \multirow{3}{*}{\textbf{Evo}}
    & MF50  & \textit{0.862} & \textbf{0.899} & 0.771 & \underline{0.895} & 0.724 & 0.734 & 0.715 & 0.342 \\
    & MP50  & 0.897 & \underline{0.926} & 0.803 & \textbf{0.932} & 0.775 & \textit{0.912} & 0.895 & 0.792 \\
    & Cross       & 0.235 & \textit{0.262} & 0.243 & \textbf{0.275} & 0.272 & \underline{0.274} & 0.227 & 0.101 \\
    \midrule
    \multirow{3}{*}{\textbf{Motif}}
    & MF50  & \textit{0.855} & \underline{0.874} & 0.779 & \textbf{0.884} & 0.767 & 0.802 & 0.716 & 0.661 \\
    & MP50  & \textit{0.850} & \underline{0.857} & 0.796 & \textbf{0.870} & 0.784 & 0.841 & 0.765 & 0.736 \\
    & Cross       & \textit{0.433} & \textbf{0.456} & 0.348 & 0.394 & 0.408 & \underline{0.441} & 0.390 & 0.329 \\
    \midrule
    \multirow{3}{*}{\textbf{Dom}}
    & MF50  & 0.634 & \underline{0.666} & 0.591 & \textbf{0.673} & 0.574 & \textit{0.642} & \textbf{--} & 0.560 \\
    & MP50  & \underline{0.645} & \underline{0.657} & 0.592 & \textbf{0.665} & 0.584 & 0.640 & \textbf{--} & 0.557 \\
    & Cross       & 0.470 & 0.506 & \textit{0.508} & 0.449 & \underline{0.525} & \textbf{0.564} & \textbf{--} & 0.468 \\
    \midrule
    \multirow{3}{*}{\textbf{BindP}}
    & MP50  & \textit{0.409}& \textbf{0.446} & 0.340 & \underline{0.421} & \textbf{--} & \textbf{--} & \textbf{--} & \textbf{--} \\
    & MP70  & \textit{0.465}& \textbf{0.494} & 0.410 & \underline{0.487} & \textbf{--} & \textbf{--} & \textbf{--} & \textbf{--} \\
    & MP90  & \textit{0.496}& \textbf{0.535} & 0.466 & \underline{0.527} & \textbf{--} & \textbf{--} & \textbf{--} & \textbf{--} \\
    \midrule
    \multirow{3}{*}{\textbf{Epi}}
    & MP50  & \textbf{0.186} & \underline{0.174} & \textit{0.169} & 0.167 & \textbf{--} & \textbf{--} & \textbf{--} & \textbf{--} \\
    & MP70  & \textit{0.184} & \underline{0.202} & 0.177 & \textbf{0.215} & \textbf{--} & \textbf{--} & \textbf{--} & \textbf{--} \\
    & MP90  & \underline{0.277} & \textbf{0.290} & 0.266 & \textit{0.270} & \textbf{--} & \textbf{--} & \textbf{--} & \textbf{--} \\
    \bottomrule
    \end{tabular}%
}
\end{table}

\subsection{Experimental Setup}

\paragraph{Model Setup}
For both residue- and fragment-level classification tasks, pretrained sequence-based models (e.g., \textsc{ESM2} \cite{lin2023esm2}, \textsc{ProtBert} \cite{elnaggar2021prottrans}) and sequence–structure models (\textsc{SaProt} \cite{su2023saprot}, \textsc{ProtSSN} \cite{tan2025protssn}) are used as frozen feature extractors. In contrast, the structure-based \textsc{GVP-GNN} \cite{gvp} is trained from scratch with all parameters updated. For residue-level tasks, the encoders output embeddings of each residue, which are passed through two linear layers with ReLU activation and dropout. For fragment-level classification, mean pooling is applied to obtain fragment representations for InterPro family prediction. In pair-level similarity evaluation, full-length sequences or fragments are encoded, mean-pooled, and compared via similarity metrics to assess family-level relationships. Parameter statistics for all models are provided in Table~\ref{tab:baseline} in Appendix~\ref{app:baseline_sec}.

\paragraph{Training Setup}

For all tasks, full-length protein sequences are truncated to a maximum of 1022 residues. Fragments are capped at 128 residues for Act, BindI, Evo, and Motif, and at 512 residues for Dom. All models are trained with a fixed random seed of 3407 to ensure reproducibility. Optimization is performed using AdamW \cite{adamw_fix} with a learning rate of 0.001 and an effective batch size of 128 via gradient accumulation. Training proceeds for up to 100 epochs, with early stopping triggered if validation performance does not improve for 10 epochs. For residue-level and fragment-level classification tasks, AUPR and accuracy on the validation set are used as early stopping criteria, respectively. All experiments are conducted on 16 NVIDIA RTX 4090D GPUs and 192 Intel(R) Xeon(R) Gold 6248R CPUs with 2 TB of memory for 45 days.

\begin{table}[t]
\caption{
Fragment-level classification performance across InterPro datasets and data splits under 50\% sequence identity. \textbf{Top-1}, \underline{Top-2}, and \textit{Top-3} results for each metric are highlighted, respectively. Detailed results are provided in Table~\ref{app:tab:fragment_cls_main_results} of Appendix~\ref{app:add_exp_sec}}
\label{tab:frag-cls}
\centering
\resizebox{\textwidth}{!}{%
    \begin{tabular}{@{}llcccccccc@{}}
    \toprule
    \multirow{2}{*}{\textbf{Target}} & \multirow{2}{*}{\textbf{Metric}} 
    & \multicolumn{4}{c}{\textbf{Sequence-only}} 
    & \multicolumn{3}{c}{\textbf{Sequence-Structure}} 
    & \multicolumn{1}{c}{\textbf{Structure-only}} \\
    \cmidrule(lr){3-6} \cmidrule(lr){7-9} \cmidrule(lr){10-10}
     & 
     & \textsc{ESM2-t30} & \textsc{ESM2-t33} & \textsc{ProtBert} & \textsc{Ankh-base} 
     & \textsc{SaProt-35M} & \textsc{SaProt-650M} & \textsc{ProtSSN} 
     & \textsc{GVP-GNN} \\
    \midrule
    \multirow{2}{*}{\textbf{Act}}
    & ACC  & 0.819 & 0.814 & 0.736 & 0.824 & \textbf{0.928} & \textbf{0.928} & \textit{0.891} & \underline{0.907} \\
    & Macro-F1  & 0.647 & 0.605 & 0.609 & 0.647 & \textit{0.807}  & \underline{0.825} & 0.764 & \textbf{0.906} \\
    \midrule
    \multirow{2}{*}{\textbf{BindI}}
    & ACC  & 0.937 & 0.934 & 0.927 & 0.920 & \underline{0.976} & \textbf{0.986} & \textit{0.972} & \textit{0.972} \\
    & Macro-F1  & \textit{0.913}& 0.753 & 0.790 & 0.718 & 0.809 & \textbf{0.957} & \underline{0.931} & 0.884 \\
    \midrule
    \multirow{2}{*}{\textbf{Evo}}
    & ACC  & 0.853 & 0.841 & 0.828 & 0.866 & \underline{0.939} & \textbf{0.950} & \textit{0.915} & 0.914 \\
    & Macro-F1  & 0.667 & 0.669 & 0.627 & 0.716 & \underline{0.849} & \textbf{0.863} & \textit{0.793} & 0.757 \\
    \midrule
    \multirow{2}{*}{\textbf{Motif}}
    & ACC  & 0.884 & \textit{0.906} & 0.884 & 0.901 & 0.901 & \textbf{0.927} & \underline{0.914} & 0.807 \\
    & Macro-F1  & 0.457 & \textit{0.542} & 0.452 & 0.499 & 0.504 & \underline{0.552} & \textbf{0.556} & 0.370 \\
    \bottomrule
    \end{tabular}%
}
\end{table}

\begin{table*}[!t]
\caption{AUC (\%) of baseline models on InterPro family alignment under two evaluation settings: \textbf{F50} (fragment-level inputs with 50\% sequence identity filtering) and \textbf{P50} (full-sequence inputs with 50\% identity). Models are grouped by modality. Cell colors indicate ranking:
 \raisebox{.3ex}{\colorbox{gold1}{\rule{0pt}{0.3em}\rule{1em}{0pt}}} Top-1, 
 \raisebox{.3ex}{\colorbox{gold2}{\rule{0pt}{0.3em}\rule{1em}{0pt}}} Top-2, 
 \raisebox{.3ex}{\colorbox{gold3}{\rule{0pt}{0.3em}\rule{1em}{0pt}}} Top-3, 
 \raisebox{.3ex}{\colorbox{gold4}{\rule{0pt}{0.3em}\rule{1em}{0pt}}} Top-4. Standard deviation over three folds is shown in parentheses.}
\label{tab:pair-cls-result}
\begin{center}
\resizebox{\linewidth}{!}{
\begin{tabular}{@{}lccccccccccc@{}}
    \toprule
    \multicolumn{2}{c}{\textbf{Model Information}} 
    & \multicolumn{2}{c}{\textbf{Act}} 
    & \multicolumn{2}{c}{\textbf{BindI}} 
    & \multicolumn{2}{c}{\textbf{Evo}} 
    & \multicolumn{2}{c}{\textbf{Motif}} 
    & \multicolumn{2}{c}{\textbf{Dom}} \\
    \cmidrule(lr){1-2} \cmidrule(lr){3-4} \cmidrule(lr){5-6} \cmidrule(lr){7-8} \cmidrule(lr){9-10} \cmidrule(lr){11-12}
    Name & Version 
    & F50 & P50 
    & F50 & P50 
    & F50 & P50 
    & F50 & P50 
    & F50 & P50 \\
    \midrule 
     \multicolumn{12}{c}{\textbf{Alignment-based Methods}} \\
    \midrule 
      \multirow{2}{*}{\textsc{Foldseek}} & 3Di       & \cellcolor{gold3} $96.0_{(0.1)}$&  \cellcolor{gold1} $96.5_{(0.2)}$& $92.6_{(0.2)}$& \cellcolor{gold3} $80.6_{(0.2)}$& \cellcolor{gold2} $88.3_{(0.1)}$& \cellcolor{gold1} $99.0_{(0.1)}$& $74.8_{(0.2)}$& $64.9_{(0.1)}$& \textbf{--} & \textbf{--} \\
        & 3Di-AA    & \cellcolor{gold2} $96.1_{(0.1)}$& \cellcolor{gold1} $96.5_{(0.2)}$& $92.6_{(0.2)}$& \cellcolor{gold4} $80.1_{(0.2)}$& \cellcolor{gold1} $88.4_{(0.2)}$& \cellcolor{gold1} $99.0_{(0.1)}$& $74.7_{(0.2)}$& $64.7_{(0.2)}$& \textbf{--} & \textbf{--} \\ 
      \textsc{TM-align}                  & mean      & $94.6_{(0.0)}$& \textbf{--} & $90.1_{(0.1)}$& \textbf{--} & \cellcolor{gold4} $67.7_{(0.1)}$ & \textbf{--} & $76.6_{(0.0)}$& \textbf{--} & \textbf{--} & \textbf{--} \\ 
      \textsc{BLAST}     & \textbf{--}         & $52.9_{(0.2)}$& $71.7_{(0.1)}$& $52.4_{(0.1)}$& $51.1_{(0.0)}$& $54.0_{(0.3)}$& \textbf{--} & $49.9_{(0.1)}$& $56.2_{(0.3)}$& \textbf{--} & \textbf{--} \\
    \midrule
       \multicolumn{12}{c}{\textbf{Sequence-only Encoder Methods}} \\
       \midrule
      \multirow{3}{*}{\textsc{ESM2}} 
        & t30 & $69.4_{(0.5)}$ & $69.2_{(0.2)}$ & $77.6_{(0.4)}$ & $65.5_{(0.2)}$ & $52.4_{(0.5)}$ & $87.5_{(0.2)}$ & $84.3_{(0.5)}$ & $68.2_{(0.3)}$ & $78.0_{(0.2)}$ & $77.4_{(0.0)}$ \\
       & t33 & $50.2_{(0.5)}$ & $70.0_{(0.3)}$ & $73.0_{(0.4)}$ & $62.3_{(0.3)}$ & $49.3_{(0.5)}$ & $89.0_{(0.2)}$ & $92.1_{(0.3)}$ & $66.1_{(0.1)}$ & $62.2_{(0.2)}$ & $66.4_{(0.1)}$ \\
       & t36 & $65.8_{(0.3)}$ & $72.9_{(0.2)}$ & $71.3_{(0.3)}$ & $67.6_{(0.4)}$ & $63.9_{(0.1)}$ & $92.1_{(0.0)}$ & $90.1_{(0.3)}$ & \cellcolor{gold2} $70.0_{(0.1)}$& $66.5_{(0.2)}$ & $66.7_{(0.0)}$ \\
       \textsc{ESM-1b} & t33 & $67.6_{(0.2)}$ & $73.8_{(0.2)}$ & $84.5_{(0.2)}$ & $69.8_{(0.2)}$ & $57.0_{(0.5)}$ & $88.4_{(0.3)}$ & $87.2_{(0.3)}$ & $58.4_{(0.4)}$ & $89.2_{(0.2)}$ & $74.7_{(0.2)}$ \\ 
      \textsc{ProtBert} & bfd & $71.4_{(0.5)}$ & $68.7_{(0.3)}$ & $84.9_{(0.4)}$ & $66.8_{(0.1)}$ & $54.6_{(0.4)}$ & $84.2_{(0.3)}$ & $85.1_{(0.2)}$ & $68.2_{(0.3)}$ & $85.3_{(0.2)}$ & $77.9_{(0.3)}$ \\ 
    \midrule 
    \multicolumn{12}{c}{\textbf{Sequence-only Encoder-Decoder Methods}} \\
       \midrule
      
      \textsc{ProtT5} & xl\_uniref50 & $91.8_{(0.1)}$ & $78.1_{(0.2)}$ & \cellcolor{gold3} $98.5_{(0.1)}$& $77.1_{(0.1)}$ & \cellcolor{gold3} $71.0_{(0.2)}$& $95.6_{(0.1)}$ & \cellcolor{gold4} $98.2_{(0.0)}$& $67.6_{(0.3)}$ & \cellcolor{gold1} $98.5_{(0.1)}$& \cellcolor{gold2}$85.1_{(0.1)}$ \\ 
      \textsc{Ankh} & base & $69.6_{(0.5)}$ & \cellcolor{gold2} $90.4_{(0.2)}$& $88.9_{(0.2)}$ & \cellcolor{gold1} $91.8_{(0.2)}$& $63.9_{(0.4)}$ & \cellcolor{gold2} $98.9_{(0.1)}$& $86.7_{(0.2)}$ & \cellcolor{gold4} $69.7_{(0.3)}$& \cellcolor{gold3} $97.6_{(0.1)}$& \cellcolor{gold1}$88.5_{(0.1)}$ \\ 
      
      \midrule
      \multicolumn{12}{c}{\textbf{Sequence-structure Methods}} \\
       \midrule
       \multirow{2}{*}{\textsc{SaProt}} & 35M\_AF2 &\cellcolor{gold4} $95.8_{(0.0)}$ & $74.6_{(0.1)}$ & $94.3_{(0.1)}$ & $71.9_{(0.2)}$ & $61.9_{(0.5)}$ & $92.7_{(0.2)}$ & $85.3_{(0.1)}$ & $66.6_{(0.2)}$ & $96.0_{(0.1)}$ & \cellcolor{gold4}$78.8_{(0.3)}$\\
       & 650M\_PDB & $82.8_{(0.2)}$ & $68.2_{(0.1)}$ & \cellcolor{gold4} $98.1_{(0.1)}$& $71.1_{(0.1)}$ & $62.6_{(0.5)}$ & $93.8_{(0.1)}$ & \cellcolor{gold2} $98.9_{(0.0)}$& $68.3_{(0.3)}$ & $91.7_{(0.1)}$ & $76.1_{(0.2)}$ \\
      \textsc{ProtSSN} & k20\_h512& $79.1_{(0.3)}$ & $64.8_{(0.2)}$ & $88.4_{(0.4)}$ & $61.2_{(0.4)}$ & $60.9_{(0.4)}$ & $86.2_{(0.1)}$ & $72.4_{(0.3)}$ & $64.0_{(0.2)}$ & $82.9_{(0.2)}$ & $69.4_{(0.0)}$ \\
      \textsc{ESM-IF} & \textbf{--} & \cellcolor{gold1} $96.5_{(0.1)}$& $70.2_{(0.2)}$ & $95.0_{(0.1)}$ & $65.6_{(0.1)}$ & $61.3_{(0.3)}$ & $90.6_{(0.4)}$ & $80.4_{(0.2)}$ & $66.0_{(0.2)}$ & \cellcolor{gold4} $97.1_{(0.2)}$& $70.5_{(0.2)}$ \\ 
      \textsc{MIF-ST} & \textbf{--} & $65.9_{(0.6)}$ & $65.9_{(0.3)}$ & $86.1_{(0.1)}$ & $59.2_{(0.3)}$ & $61.3_{(0.3)}$ & $80.3_{(0.2)}$ & $50.2_{(0.4)}$ & $66.3_{(0.6)}$ & $78.6_{(0.2)}$ & $66.7_{(0.0)}$ \\
      \textsc{TM-vec} & swiss\_large & $93.6_{(0.2)}$ & \cellcolor{gold3} $89.9_{(0.2)}$& \cellcolor{gold2} $98.6_{(0.0)}$& \cellcolor{gold2} $82.4_{(0.0)}$& $67.4_{(0.2)}$ & \cellcolor{gold4} $96.2_{(0.1)}$& \cellcolor{gold1} $99.4_{(0.0)}$& \cellcolor{gold1} $71.7_{(0.3)}$& \cellcolor{gold2} $98.2_{(0.1)}$& $59.9_{(0.2)}$ \\
      \textsc{ProstT5} & AA2fold & $90.8_{(0.1)}$ & \cellcolor{gold4} $80.7_{(0.3)}$& \cellcolor{gold1} $99.5_{(0.0)}$& $79.2_{(0.0)}$ & $55.6_{(0.5)}$ & \cellcolor{gold3} $98.2_{(0.0)}$& \cellcolor{gold3} $98.5_{(0.0)}$& \cellcolor{gold3} $69.8_{(0.2)}$& \cellcolor{gold1} $98.5_{(0.1)}$& \cellcolor{gold3}$79.3_{(0.2)}$ \\
    \bottomrule
\end{tabular}
}
\end{center}
\end{table*}

\subsection{Evaluated Methods}
The summary of all baseline models, including architecture type, version, task scope, parameter count, and implementation source, is presented in Table~\ref{tab:baseline} in Appendix.

\paragraph{Classification Baselines}
For the residue-level and fragment-level classification tasks, we evaluate a set of pretrained models spanning diverse architectures. These include sequence-based language models such as \textsc{ESM2} \cite{lin2023esm2}, \textsc{ProtBert} \cite{elnaggar2021prottrans}, and \textsc{Ankh} \cite{elnaggar2023ankh}; sequence–structure hybrid models such as \textsc{SaProt} \cite{su2023saprot} and \textsc{ProtSSN} \cite{tan2025protssn}; and a structure-only geometric network, \textsc{GVP-GNN} \cite{gvp}.
\paragraph{Similarity Scoring Baselines}
For the pair-wise similarity evaluation task, we extract mean embeddings from pretrained language models, including \textsc{ESM2} \cite{lin2023esm2}, \textsc{ProtBert} \cite{elnaggar2021prottrans}, \textsc{ESM-1b} \cite{rives2021esm1b}, \textsc{ProtT5} \cite{elnaggar2021prottrans}, \textsc{Ankh} \cite{elnaggar2023ankh}, \textsc{TM-vec} \cite{hamamsy2024tmvec}, and two inverse folding models, \textsc{MIS-ST} \cite{mifst} and \textsc{ESM-IF1} \cite{hsu2022esm-if1}. For alignment-based baselines, we include \textsc{BLAST} \cite{altschul1990blast} (sequence alignment), \textsc{TM-align} \cite{zhang2005tmalign} (structure alignment), and \textsc{Foldseek} \cite{van2024foldseek}, which supports both structure-only (3Di) and joint sequence-structure (3Di-AA) comparisons.

\subsection{Residue-level Binary Calssification}
Table~\ref{tab:residue-cls} lists AUPR on seven residue-annotation benchmarks under mixed-family (in-distribution) and cross-family (out-of-distribution) splits. Some experiments are omitted due to missing structural inputs or prohibitive computational costs. Key observations are:
\begin{itemize}[leftmargin=*]
    \item \textbf{Language models perform strongly on in-distribution splits.} \textsc{Ankh-base} attains the highest AUPR on 7/15 InterPro Mix tasks, while \textsc{ESM2-t33} dominates the \textit{BindP} and \textit{Epi} mixes, confirming that sequence is sufficient when test proteins remain close to the training set.
    \item \textbf{Sequence–structure models generalize better for unseen families.} \textsc{SaProt-650M} achieves the best or second-best performance across all InterPro Cross splits, and shows a notable advantage in domain-level classification (e.g., +5.6\% AUPR over \textsc{ProtBert}).
    \item \textbf{Cross-family residue prediction remains highly challenging.} On \textit{Act} and \textit{BindI} the best AUPR plummets by 70–80\% in Cross, whereas \textit{Dom} drops by $<10$\%, suggesting that catalytic and binding residues are harder to extrapolate than domain-wide patterns.
    \item \textbf{Dataset properties strongly affect difficulty.} Mix splits from InterPro are relatively well-structured and easier to predict. In contrast, \textit{Epi} remains extremely difficult—no model achieves an AUPR above 0.3 across all sequence identity levels.
\end{itemize}

\subsection{Fragment-level Multi-Class Classification}
Table \ref{tab:frag-cls} reports ACC and Macro-F1 on four InterPro targets (Act, BindI, Evo, Motif) with 50\% sequence-identity filtering.

\begin{itemize}[leftmargin=*]
    \item \textbf{Sequence-structure models are consistently superior.} \textsc{SaProt-650M} or \textsc{ProtSSN} ranks first on 7/8 metrics; e.g.\ \textsc{SaProt-650M} outperforms the strongest protein language model (\textsc{ESM2-t33}) on \textit{Act} by +11.4\% ACC and +22\% Macro-F1.
    \item \textbf{Structure-only models show strong task-specific performance.} \textsc{GVP-GNN} matches alignment-aware models on \textit{Act} and \textit{BindI} (Macro-F1 = 0.906 / 0.884) but lags on \textit{Motif}, indicating pure structure is task-dependent.
    \item \textbf{Sequence models are more sensitive to class imbalance.} While ACC exceeds 80\%, their Macro-F1 is 15–20\% lower; sequence-structure models cut this gap to $\sim$10\%, showing better robustness to skewed label distributions.
\end{itemize}

\subsection{Pairwise Functional Similarity Scoring}

Table \ref{tab:pair-cls-result} reports AUC (\%) for family-level alignment under two low-identity conditions—\textbf{F50} (fragment inputs clustered at 50\% identity) and \textbf{P50} (protein inputs at 50\% identity). Because exhaustive structural alignment (e.g., \textsc{TM-align}) is prohibitively expensive at this scale, several entries are left blank; baseline models description and full command lines are given in Appendix~\ref{app:alignment_sec}.

\begin{itemize}[leftmargin=*]
    \item \textbf{Structure-based aligners remain the gold standard.} \textsc{Foldseek} delivers near-perfect performance, topping 3/8 settings and peaking at 99.0\% AUC on \textit{Evo\_P50}. \textsc{TM-align} is likewise competitive when evaluated, whereas sequence-only \textsc{BLAST} trails by $>40\%$ on every task, underscoring the advantage of structural information.
    \item \textbf{Large encoder–decoder models close much of the gap.}  \textsc{ProtT5} attains 98.5\% on \textit{BindI\_F50} and 98.2\% on \textit{Motif\_F50}, outperforming all pure sequence encoders (\textsc{ESM2}, \textsc{ESM-1b}, \textsc{ProtBert}) by 7–20\% and surpassing \textsc{TM-align} on three fragment settings. Ankh shows similar strength on full‐sequence inputs (\textit{Act\_P50}: 90.4\% vs.\ 69–74\% for vanilla encoders).
    \item \textbf{Sequence–structure hybrids are highly alignment-aware.}  \textsc{TM-vec} reaches top-2 ranking in 5 of 10 cells (e.g., 99.4\% on \textit{Motif\_P50}; 98.2\% on \textit{Dom\_P50}), while \textsc{SaProt-650M} attains 98.1\% on \textit{BindI\_P50}. These results indicate that injecting structural inductive bias enables LM embeddings to rival specialised aligners at a fraction of the computational cost.

\end{itemize}

\section{Related Work}
\paragraph{Protein-wise Tasks} A variety of benchmarks have been developed to support machine learning on protein sequence and structure data. Early efforts such as TAPE \cite{rao2019tape} and ProteinNet \cite{alquraishi2019proteinnet} focused on sequence-level tasks including secondary structure prediction, contact prediction, and remote homology classification. More recently, benchmarks like PEER \cite{xu2022peer}, PETA \cite{tan2023peta}, VenusFactory \cite{tan2025venusfactory}, and ProteinGLUE \cite{capel2022proteinglue} introduced multi-task evaluations for protein sequence understanding, emphasizing sequence-level predictions across diverse annotation types \cite{almagro2017deeploc,Khurana2018deepsol}. Envision \cite{gray2018envision}, DeepSequence \cite{riesselman2018deepsequence}, and ProteinGym \cite{notin2024proteingym} advanced large-scale evaluation for fitness prediction under zero-shot or supervised mode, while FLIP \cite{dallago2021flip} curated different split strategies (e.g., one-vs-rest, which trains models on single mutations and tests on the rest of the high-order mutations) to cover various scenarios. ProteinShake \cite{kucera2023proteinshake} standardized structural datasets and task formulations across graph, point cloud, and voxel-based representations for protein structures. 

\paragraph{Protein-pair Tasks} Protein-pair modeling tasks encompass a broad spectrum of physical and functional interactions. PEER \cite{xu2022peer} includes interaction classification in human \cite{pan2010human_ppi} and yeast \cite{guo2008yeast_ppi} PPI networks, as well as affinity regression using SKEMPI \cite{jankauskaite2019skempi2}. Structural datasets such as MaSIF \cite{gainza2020masif} and DIPS-plus \cite{morehead2023dips} provide high-quality annotations of protein–protein interfaces, enabling geometric modeling of interaction surfaces. The recent HDPL pocketome \cite{moine2024hdpl} expands this scope by offering pocket-centric structural data related to PPIs and PPI-related ligand binding sites. Functional networks like STRING \cite{szklarczyk2019string} support large-scale classification of biological associations. In the protein–ligand domain, PDBbind \cite{wang2005pdbbind} and the Ligand Binding Affinity (LBA) tasks in Atom3D \cite{townshend2020atom3d} provide affinity labels derived from co-crystal structures. While these resources facilitate pairwise prediction and interaction modeling, they generally lack residue-level supervision, limiting their utility in evaluating fine-grained functional inference.

\section{Disscusion and Conclusion}\label{sec:dis_con}
This work presents \venus, a comprehensive benchmark for evaluating deep learning models in protein representation learning, with a specific focusing on functional classification and similarity scoring at residue, fragment, and protein levels. To the best of our knowledge, it is the first and largest benchmark designed to assess model understanding of protein function at such fine granularity, comprising over 878k samples and 56 datasets. Unlike existing benchmarks that focus on coarse protein-level annotations, \venus~enables more precise and biologically relevant evaluation by capturing fine-scale functional signals.

As deep learning models for proteins continue to grow in number and complexity, there is an increasing need for biologically meaningful benchmarks that involve careful data curation and cleaning, and that reflect application-driven learning and evaluation objectives. \venus~fulfills this need by offering diverse tasks, standardized splits, and consistent evaluation metrics, enabling fair and informative comparisons across models. We expect \venus~to play a central role in advancing functional protein modeling, supporting both methodological progress in representation learning and practical applications such as enzyme mining, drug target design, and structural proteomics.

\section*{Acknowledgements}
This work was supported by the National Science Foundation of China (Grant Number 62302291), National Key Research and Development Program of China (2024YFA0917603), Computational Biology Key Program of Shanghai Science and Technology Commission (23JS1400600), and Science and Technology Innovation Key R\&D Program of Chongqing (CSTB2022TIAD-STX0017).

% \section*{References}
\bibliography{2reference}
\bibliographystyle{unsrt}

\newpage
\appendix
\section{Limitations and Broader Impact}\label{app:limitation_impact_sec}

\paragraph{Limitations.}
While our benchmark provides a biologically grounded and fine-grained evaluation suite across residues, fragments, and pairwise similarity, several limitations remain. First, although we include a range of representative baselines, some recent state-of-the-art models—especially those requiring specialized resources—are not yet evaluated. Second, the current tasks focus on subprotein-level understanding and do not fully cover broader settings like full-sequence function prediction or structure-based inference. Third, we primarily adopt standard evaluation metrics such as AUPR, accuracy, and F1, which may overlook aspects like robustness or calibration. Despite filtering and preprocessing, the datasets may still contain biases or distributional imbalances. In addition, certain experiments were skipped due to lack of structure availability or excessive runtime.

\paragraph{Broader Impact.}
This benchmark is designed to support progress in protein representation learning by offering a standardized, interpretable, and reproducible platform. By releasing all data and code, we aim to lower the barrier to entry, encourage fair comparison, and foster community collaboration. However, as with any benchmark, there is a risk of overfitting to predefined tasks or metrics. Moreover, if models trained or evaluated using this benchmark are applied in sensitive contexts such as drug discovery or clinical diagnostics, careful consideration of ethical and safety implications is necessary. We encourage responsible use and welcome contributions to extend the benchmark’s coverage and long-term utility.

\section{Dataset}
% Technical appendices with additional results, figures, graphs and proofs may be submitted with the paper submission before the full submission deadline (see above), or as a separate PDF in the ZIP file below before the supplementary material deadline. There is no page limit for the technical appendices.

\subsection{Sequence Length Distribution}
\begin{figure}[ht]
    \centering
    \includegraphics[width=\textwidth]{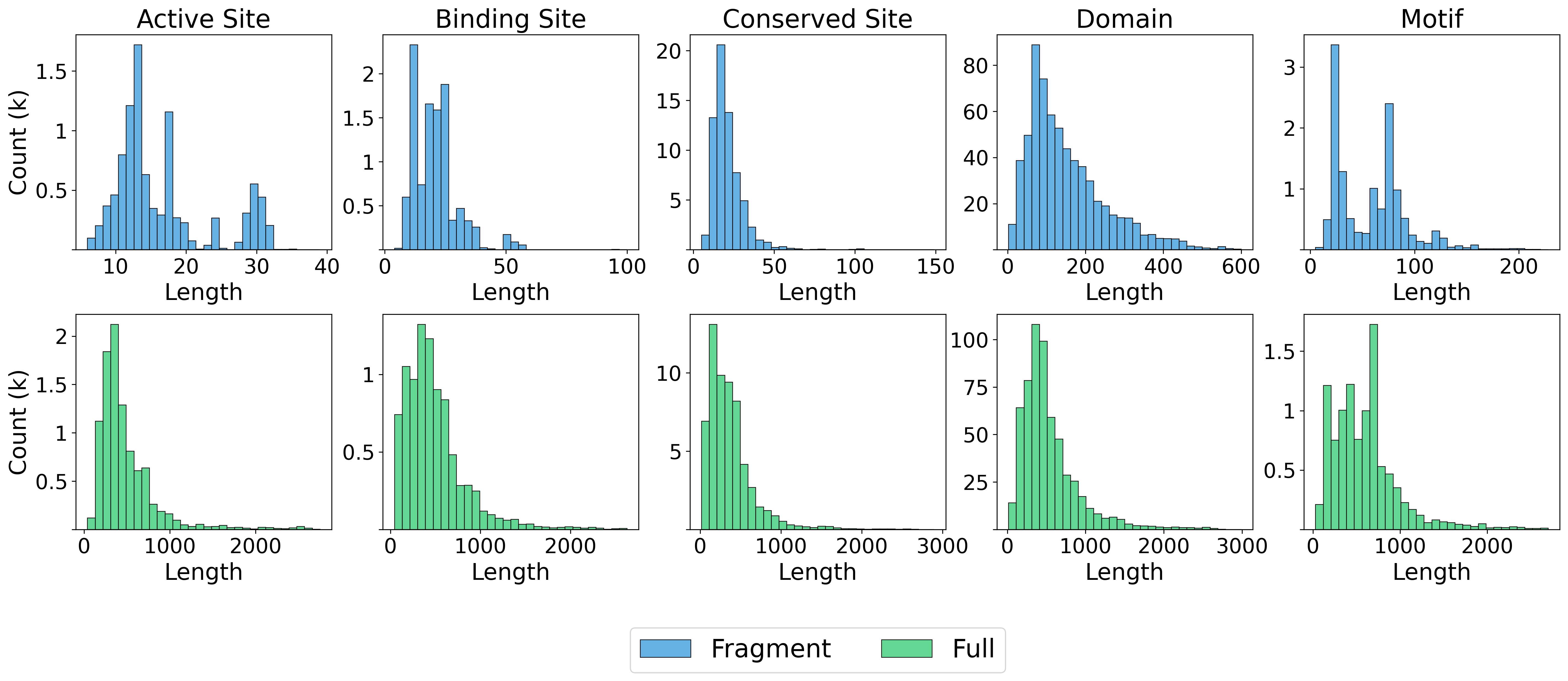}
    \vspace{-1mm}
    \caption{Sequence length distribution of the \venus~InterPro benchmark.}
    \label{fig:seq_dis}
\end{figure}

\begin{table}[ht]
\centering
\caption{Summary of fragment and full-sequence length statistics for each InterPro category.}
\label{tab:seq-length-stats}
\resizebox{0.6\textwidth}{!}{
    \begin{tabular}{lcccccc}
    \toprule
    \textbf{Target} & \textbf{Type} & \textbf{\#Samples} & \textbf{Min} & \textbf{Max} & \textbf{Mean} & \textbf{Median} \\
    \midrule
    \multirow{2}{*}{\textbf{Act}} 
    & Fragment     & 9,767  & 6   & 39   & 16.54  & 14.00 \\
    & Protein    & 9,667  & 39  & 2,753 & 482.51 & 379.00 \\
    \midrule
    \multirow{2}{*}{\textbf{BindI}} 
    & Fragment     & 10,562 & 4   & 100  & 20.75  & 19.00 \\
    & Protein    & 8,959  & 45  & 2,631 & 486.54 & 415.00 \\
    \midrule
    \multirow{2}{*}{\textbf{Evo}} 
    & Fragment     & 66,916 & 5   & 149  & 20.93  & 19.00 \\
    & Protein    & 59,948 & 16  & 2,896 & 365.88 & 305.00 \\
    \midrule
    \multirow{2}{*}{\textbf{Dom}} 
    & Fragment     & 653,259 & 2   & 600  & 150.54 & 124.00 \\
    & Protein    & 595,443 & 16  & 2,988 & 537.41 & 444.00 \\
    \midrule
    \multirow{2}{*}{\textbf{Motif}} 
    & Fragment     & 13,137 & 5   & 228  & 57.02  & 62.00 \\
    & Protein    & 10,271 & 30  & 2,699 & 595.15 & 558.00 \\
    \bottomrule
    \end{tabular}
}
\end{table}

Figure \ref{fig:seq_dis} and Table \ref{tab:seq-length-stats} show the length distributions of annotated protein fragments (top) and their corresponding full-length protein sequences (bottom) across five InterPro categories. To facilitate visualization, outliers were excluded: domain fragments longer than 600 residues, motif fragments exceeding 230 residues, and full-length proteins longer than 3000 residues.

Several trends can be observed:
\begin{itemize}
    \item At the fragment level, \textbf{Act}, \textbf{BindI}, and \textbf{Evo} sites exhibit relatively short lengths, typically under 50 residues, with \textbf{Act} sites showing clear multimodal peaks due to specific catalytic motifs. \textbf{Dom} and \textbf{Motif} fragments show broader distributions, with domain fragments exhibiting a long tail.
    \item In contrast, full-length proteins follow a typical long-tailed distribution across all categories, with most proteins under 1000 residues but a small number extending beyond 2000. The distributions are highly skewed, especially in the \textbf{Dom} and \textbf{Evo} datasets, reflecting the diversity of protein sizes.
\end{itemize}
These distributions motivate the design of separate fragment- and full-sequence benchmarks, as the input length significantly impacts model performance and scalability.

% \newpage
\subsection{InterPro Label Distribution}

\begin{figure}[ht]
    \centering
    \includegraphics[width=\textwidth]{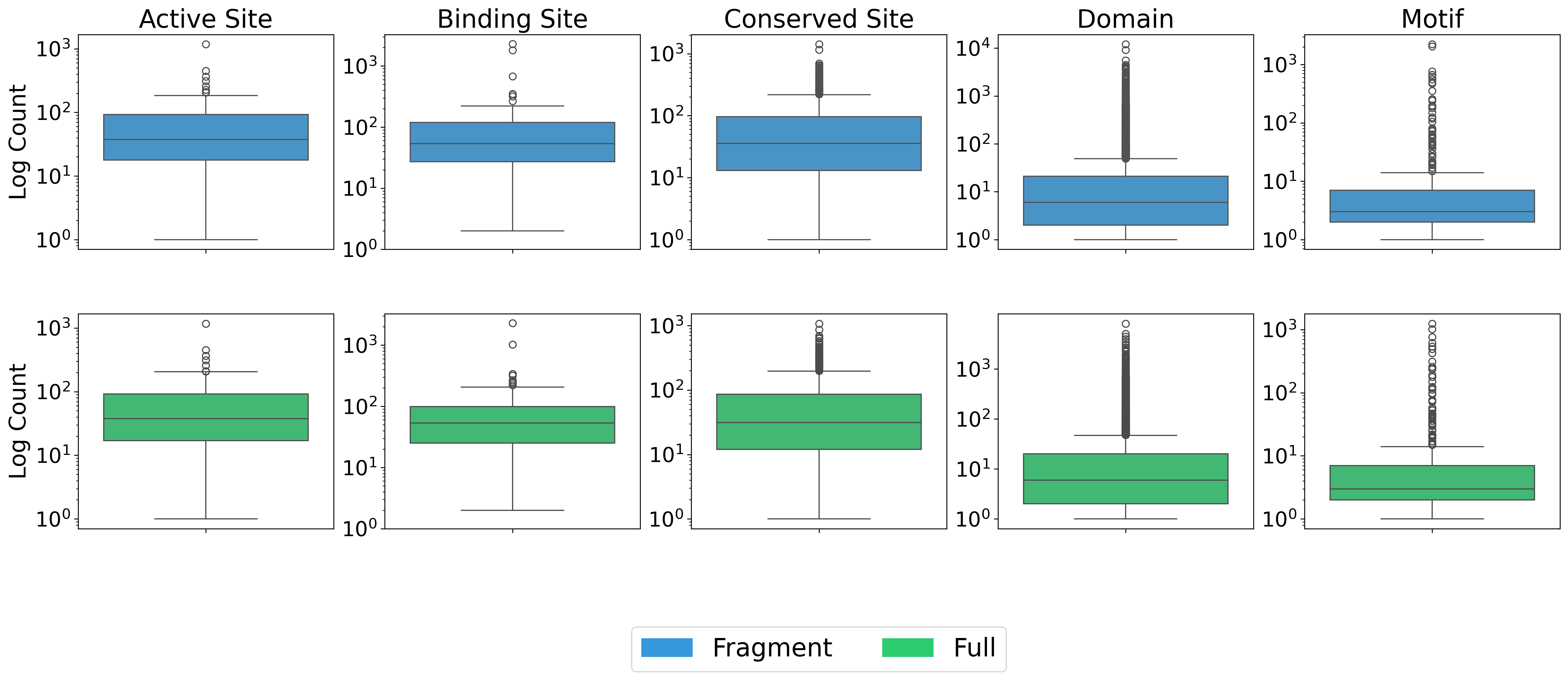}
    \vspace{-1mm}
    \caption{InterPro label distribution of the \venus~ benchmark.}
    \label{fig:label-distribution-box}
\end{figure}

\begin{table}[ht]
\centering
\caption{Statistics of InterPro type label frequency across five categories, computed separately on annotated fragments and full sequences.}
\label{tab:interpro-label-stats}
\resizebox{0.6\textwidth}{!}{
    \begin{tabular}{lcccccc}
    \toprule
    \textbf{Target} & \textbf{Type} & \textbf{\#Types} & \textbf{Min} & \textbf{Max} & \textbf{Mean} & \textbf{Median} \\
    \midrule
    \multirow{2}{*}{\textbf{Act}}
    & Fragment     & 132   & 1   & 1176 & 73.99  & 37.50 \\
    & Protein    & 132   & 1   & 1172 & 73.23  & 37.50 \\
    \midrule
    \multirow{2}{*}{\textbf{BindI}}
    & Fragment     & 76    & 2   & 2293 & 138.97 & 53.50 \\
    & Protein    & 76    & 2   & 2279 & 117.88 & 53.00 \\
    \midrule
    \multirow{2}{*}{\textbf{Evo}}
    & Fragment     & 740   & 1   & 1418 & 90.43  & 36.00 \\
    & Protein    & 740   & 1   & 1074 & 81.01  & 31.50 \\
    \midrule
    \multirow{2}{*}{\textbf{Dom}}
    & Fragment     & 12,529 & 1  & 12002 & 52.14 & 6.00 \\
    & Protein    & 12,580 & 1  & 8109  & 47.33 & 6.00 \\
    \midrule
    \multirow{2}{*}{\textbf{Motif}}
    & Fragment     & 440   & 1   & 2222 & 29.86  & 3.00 \\
    & Protein    & 454   & 1   & 1247 & 22.62  & 3.00 \\
    \bottomrule
    \end{tabular}
}
\end{table}

Based on Table~\ref{tab:interpro-label-stats} and Figure~\ref{fig:label-distribution-box}, we analyze the distribution of residue-level labels across InterPro categories. Due to the extreme imbalance in class sizes, we apply a log-scale transformation when visualizing the number of annotations per InterPro type.

\begin{itemize}
    \item \textbf{Severe long-tail distribution across datasets.} All five datasets exhibit substantial class imbalance. For example, the domain task has over 12,000 InterPro types, but a median of only 6 annotated proteins per type at both the fragment and full-sequence level. Similarly, motif labels have a median count of 3, emphasizing the prevalence of rare classes.
    \item \textbf{Skewed distribution dominated by few frequent families.} In most datasets, a small number of InterPro types contribute a disproportionate number of samples. For instance, the top 5 types in the binding site fragments account for over 5,000 samples, while many other types appear fewer than 10 times.
    \item \textbf{Fragments exhibit slightly denser annotation than full sequences.} Across all datasets, the mean and median counts per InterPro type are consistently higher at the fragment level than for full sequences. This suggests fragments are more focused on annotated functional regions, whereas full sequences dilute sparse labels across longer chains.
\end{itemize}
These trends reinforce the importance of using macro-averaged metrics and highlight the difficulty of learning under class-imbalanced, fine-grained label regimes—especially for tasks such as domain and motif classification.

\subsection{Dataset Numerical Split Detail}\label{app:split_detail}

\begin{table}[ht]
\caption{Residue-level: Number of train/validation/test examples under mixed-family (Mix50) and cross-family splits. “Fragment” and “Protein” refer to clustering at the fragment and full-sequence levels, respectively.}
\label{app:tab:residue-split-50}
\centering
\resizebox{\textwidth}{!}{%
    \begin{tabular}{@{}llcccc@{}}
    \toprule
    \multirow{2}{*}{\textbf{Target}} & \multirow{2}{*}{\textbf{Data Source}} & \multicolumn{2}{c}{\textbf{\# Train/Validation/Test (Mix50)}} & \multicolumn{2}{c}{\textbf{\# Train/Validation/Test (Cross)}} \\ \cmidrule(l){3-4} \cmidrule(l){5-6}
     &  & \multicolumn{1}{c}{Fragment} & \multicolumn{1}{c}{Protein} & \multicolumn{1}{c}{Family} & \multicolumn{1}{c}{Protein} \\ \midrule
    \textbf{Act} & InterPro \cite{paysan2023interpro} & 1,488/186/186 & 2,929/366/367 & 104/14/14 & 7,701/880/1086 \\
    \textbf{BindI} & InterPro \cite{paysan2023interpro} & 1,640/205/205 & 2,366/296/296 & 60/8/8 & 7,729/551/679 \\     
    \textbf{BindP} & BioLiP \cite{yang2012biolip} & \textbf{--} & 19,412/2,426/2,427 & \textbf{--} & \textbf{--} \\
    \textbf{Evo} & InterPro \cite{paysan2023interpro} & 10,383/1,298/1,298 & 13,552/1,694/1,694 & 592/74/74 & 48,437/5,445/6,006 \\
    \textbf{Motif} & InterPro \cite{paysan2023interpro} & 2,008/251/251 & 2,720/340/341 & 362/46/46 & 7,799/1,045/1,427 \\
    \textbf{Dom} & InterPro \cite{paysan2023interpro} & 84,489/10,561/10,562 & 113,607/14,201/14,201 & 10,065/1,258/1,260 & 477,149/56,600/61,705 \\
    \textbf{Epi} & SAbDab \cite{dunbar2014sabdab} & \textbf{--} & 828/103/104 & \textbf{--} & \textbf{--} \\ \bottomrule
    \end{tabular}%
}
\end{table}

\begin{table}[ht]
\caption{Residue-level: Number of train/validation/test examples under mixed-family at 70\% and 90\% sequence identity. “Fragment” and “Protein” refer to clustering at the fragment and protein levels.}
\label{app:tab:residue-split-70-90}
\centering
\resizebox{\textwidth}{!}{%
    \begin{tabular}{@{}llcccc@{}}
    \toprule
    \multirow{2}{*}{\textbf{Target}} & \multirow{2}{*}{\textbf{Data Source}} & \multicolumn{2}{c}{\textbf{\# Train/Validation/Test (Mix70)}} & \multicolumn{2}{c}{\textbf{\# Train/Validation/Test (Mix90)}} \\ \cmidrule(l){3-4} \cmidrule(l){5-6}
     &  & \multicolumn{1}{c}{Fragment} & \multicolumn{1}{c}{Protein} & \multicolumn{1}{c}{Fragment} & \multicolumn{1}{c}{Protein} \\ \midrule
    \textbf{Act} & InterPro \cite{paysan2023interpro} & 2,724/340/341 & 5,378/672/673 & 5,269/659/659 & 7,279/910/910 \\
    \textbf{BindI} & InterPro \cite{paysan2023interpro} & 3,016/377/377 & 4,432/554/555 & 5,306/663/664 & 6,428/803/804 \\     
    \textbf{BindP} & BioLiP \cite{yang2012biolip} & \textbf{--} & 25,137/3,142/3,142 & \textbf{--} & 32,394/4,049/4,050 \\
    \textbf{Evo} & InterPro \cite{paysan2023interpro} & 17,459/2,182/2,183 & 27,848/3,481/3,481 & 33,636/4,205/4,205 & 42,869/5,359/5,359 \\
    \textbf{Motif} & InterPro \cite{paysan2023interpro} & 3,539/442/443 & 4,624/578/579 & 5,472/684/685 & 6,715/839/840 \\
    \textbf{Dom} & InterPro \cite{paysan2023interpro} & 160,201/20,025/20,026 & 215,652/26,957/26,957 & 263,870/32,984/32,984 & 348,944/43,618/43,618 \\
    \textbf{Epi} & SAbDab \cite{dunbar2014sabdab} & \textbf{--} & 996/124/125 & \textbf{--} & 1,524/190/191 \\ \bottomrule
    \end{tabular}%
}
\end{table}

\begin{table}[ht]
\caption{Fragment-level: Number of train/validation/test examples under mixed-family at 50\%, 70\%, and 90\% sequence identity clustering at the fragment level.}
\label{app:tab:fragment-split-50-70-90}
\centering
\resizebox{0.7\textwidth}{!}{%
    \begin{tabular}{@{}lccc@{}}
    \toprule
    \multirow{2}{*}{\textbf{Target}} &
    \multicolumn{3}{c}{\textbf{\# Train/Validation/Test}} \\
    \cmidrule{2-4}
    & MF50 & MF70 & MF90 \\
    \midrule
    \textbf{Act} & 1,545/191/193 & 2,777/352/358 & 5,344/670/670 \\
    \textbf{BindI} & 2,558/294/287 & 4,075/511/472 & 6,487/830/826 \\     
    \textbf{Evo} & 12,880/1,596/1,613 & 21,140/2,604/2,596 & 38,736/4,910/4,843 \\
    \textbf{Motif} & 3,083/362/372 & 5,123/630/589 & 7,516/949/928 \\
    \textbf{Dom} & 105,110/13,112/13,011 & 188,870/23,565/23,762 & 300,661/37,539/37,742 \\
    \bottomrule
    \end{tabular}%
}
\end{table}

\begin{table}[ht]
\caption{Detailed number information of the unsupervised pair similarity evaluation task. “\# Protein-P/N” and “\# Frag-P/N” denote the total number of positive and negative pairs sampled by Protein sequences or fragments within InterPro families. “\# Protein/Frag-pdb” denotes whether the structures of protein sequences or fragments are available.}
\centering
\label{app:tab:unsup-task-detail}
\resizebox{\textwidth}{!}{%
    \begin{tabular}{llcccccc}
    \hline
    \textbf{Target} & \textbf{Data Source} & \textbf{\# Protein-P} & \textbf{\# Protein-N} & \textbf{\# Frag-P} & \textbf{\# Frag-N} & \textbf{Protein-pdb} & \textbf{Frag-pdb} \\ \hline
    \textbf{Act}   & InterPro \cite{paysan2023interpro}      & 1,314,757          & 45,405,854         & 1,331,749          & 46,360,512   & \checkmark &    \checkmark  \\
    \textbf{BindI}   & InterPro \cite{paysan2023interpro}     & 3,550,288          & 36,577,073         & 4,957,073          & 50,815,568     & \checkmark &    \checkmark    \\
    \textbf{Evo}  & InterPro \cite{paysan2023interpro}    & 7,710,941          & 1,789,140,437      & 9,990,111          & 2,228,851,959  & \checkmark &    \checkmark    \\
    \textbf{Motif}   & InterPro \cite{paysan2023interpro}       & 2,415,212          & 50,326,373         & 5,962,485          & 81,732,661     & \checkmark &    \checkmark    \\
    \textbf{Dom}   & InterPro \cite{paysan2023interpro}      & 217,681,629        & 177,064,753,702    & 346,047,386        & 215,260,712,060  & \checkmark &    \checkmark  \\ \hline
    \end{tabular}
}
\end{table}

\paragraph{Pre-filtering and Clustering.} For all InterPro-based datasets, we apply a pre-filtering step to remove sequences lacking predicted structures from the AlphaFold Protein Structure Database \cite{varadi2022alphafolddb}, ensuring structural consistency for downstream evaluations. Following this, we perform sequence identity clustering using MMseqs2 \cite{steinegger2017mmseqs2} under varying identity thresholds (50\%, 70\%, and 90\%) to construct non-redundant splits at both the fragment and full-sequence levels. Clustering is conducted with a coverage mode of 1 (query coverage), and a minimum coverage of 0.8.

\paragraph{Residue-level Split Settings.} Tables~\ref{app:tab:residue-split-50} and~\ref{app:tab:residue-split-70-90} summarize the number of train/validation/test examples under different split strategies. Table~\ref{app:tab:residue-split-50} reports counts under the mixed-family (Mix50) and cross-family splits, where fragment-level and full-sequence clustering are applied separately. Table~\ref{app:tab:residue-split-70-90} further breaks down the mixed-family splits at 70\% and 90\% sequence identity thresholds. InterPro-based datasets support all three types of splits, while BioLiP and SAbDab only provide full-sequence annotations and thus are limited to full-protein splits. \textbf{Dom} and conserved datasets are the largest in scale, enabling more comprehensive evaluations across clustering thresholds.

\paragraph{Fragment-level Split Settings.} 
Table~\ref{app:tab:fragment-split-50-70-90} presents the number of train/validation/test examples across five InterPro targets under mixed-family splits with increasing sequence identity thresholds (50\%, 70\%, and 90\%) applied at the fragment level. As expected, raising the identity threshold increases the number of retained fragments, approximately doubling the dataset size from MF50 to MF90. For instance, \textbf{Act} expands from 1.5k to 5.3k training fragments, while \textbf{Dom} scales from 105k to over 300k. This progression supports finer-grained control over redundancy and task difficulty, enabling evaluation across a spectrum of local similarity conditions. The setting facilitates analysis of model robustness to fragment diversity and homologous signal dilution.

\paragraph{Pair-level Statistics.}
Table~\ref{app:tab:unsup-task-detail} reports the number of positive and negative pairs for unsupervised similarity evaluation. Positive pairs share the same InterPro family, while negatives are drawn from different families. All tasks exhibit a strong imbalance, especially in large-scale domains (e.g., over 177 billion negative pairs). Structural coverage remains high across both full sequences and fragments, enabling comprehensive evaluation under both sequence- and structure-based settings.

\section{Baselines}\label{app:baseline_sec}

\begin{table}[ht]
\caption{Summary of baseline models by input modality. “Task” indicates evaluation scope: “All” denotes all three tasks, “Sup.” refers to supervised classification tasks only, and “Pair” to the pairwise functional similarity scoring task. We report model type, version, parameters, embedding size, and implementation source (via Hugging Face, GitHub, or Conda).}
\label{tab:baseline}
\centering
\resizebox{\textwidth}{!}{%
\begin{tabular}{@{}llcccccl@{}}
\toprule
\textbf{Input} & \textbf{Model} & \textbf{Version} & \textbf{Task} & \textbf{\# Params} & \textbf{\# Train. Params} & \textbf{Embed. Dim} & \textbf{Implementation} \\ \midrule
\multirow{10}{*}{Sequence-Only} 
  & \multirow{3}{*}{\textsc{ESM2} \cite{lin2023esm2}} 
  & t30 & All & 150M & 410K & 640 & \href{https://huggingface.co/facebook/esm2_t30_150M_UR50D}{HF: ESM2-t30} \\
  &   & t33 & All & 652M & 1.6M & 1,280 & \href{https://huggingface.co/facebook/esm2_t33_650M_UR50D}{HF: ESM2-t33} \\
  &   & t36 & Pair. & 3,000M & \textbf{--} & 2,560 & \href{https://huggingface.co/facebook/esm2_t36_3B_UR50D}{HF: ESM2-t36} \\ 
  & \textsc{ESM-1b} \cite{rives2021esm1b} & t33 & Pair. & 652M & \textbf{--} & 1,280 & \href{https://huggingface.co/facebook/esm1b_t33_650M_UR50S}{HF: ESM-1b} \\
  & \textsc{ProtBert} \cite{elnaggar2021prottrans} & uniref & All & 420M & 1.0M & 1,024 & \href{https://huggingface.co/Rostlab/prot_bert_bfd}{HF: ProtBert} \\
  & \textsc{ProtT5} \cite{elnaggar2021prottrans} & xl\_uniref50 & Pair. & 3,000M & \textbf{--} & 1,024 & \href{https://huggingface.co/Rostlab/prot_t5_xl_uniref50}{HF: ProtT5} \\
  & \textsc{Ankh} \cite{elnaggar2023ankh} & base & All & 450M & 591K & 768 & \href{https://huggingface.co/ElnaggarLab/ankh-base}{HF: Ankh} \\ 
  & \textsc{TM-vec} \cite{hamamsy2024tmvec} & swiss\_large & Pair. & 3,034M & \textbf{--} & 512 & \href{https://github.com/tymor22/tm-vec}{Github: TM-vec} \\
  & \textsc{ProstT5} \cite{heinzinger2023prostt5} & AA2fold & Pair. & 3,000M & \textbf{--} & 1024 & \href{https://huggingface.co/Rostlab/ProstT5}{HF: ProstT5} \\
  & \textsc{BLAST} \cite{altschul1990blast} & \textbf{--} & Pair. & \textbf{--} & \textbf{--} & \textbf{--} & \href{https://anaconda.org/bioconda/blast}{Conda: BLAST} \\
  \midrule

\multirow{6}{*}{Sequence-Structure} 
  & \multirow{2}{*}{\textsc{SaProt} \cite{su2023saprot}} 
    & 35M\_AF2 & All & 35M & 231K & 480 & \href{https://huggingface.co/westlake-repl/SaProt_35M_AF2}{HF: SaProt-AF2} \\
  &   & 650M\_PDB & All & 650M & 1.6M & 1,280 & \href{https://huggingface.co/westlake-repl/SaProt_650M_PDB}{HF: SaProt-PDB} \\ 
  & \textsc{ProtSSN} \cite{tan2025protssn} & k20\_h512 & All & 800M & 1.6M & 1,280 & \href{https://huggingface.co/ai4protein/ProtSSN}{HF: ProtSSN} \\
  & \textsc{ESM-IF1} \cite{hsu2022esm-if1} & \textbf{--} & Pair. & 148M & \textbf{--} & 512 & \href{https://huggingface.co/katielink/esm_if1_gvp4_t16_142M_UR50}{HF: ESM-IF1} \\
  & \textsc{MIS-ST} \cite{mifst} & \textbf{--} & Pair. & 643M & \textbf{--} & 256 & \href{https://github.com/microsoft/protein-sequence-models}{Github: MIF-ST} \\
  & \textsc{Foldseek} \cite{van2024foldseek} & 3Di-AA & Pair. & \textbf{--} & \textbf{--} & \textbf{--} & \href{https://anaconda.org/bioconda/foldseek}{Conda: Foldseek} \\
  \midrule

\multirow{3}{*}{Structure-Only} 
  & \textsc{GVP-GNN} \cite{gvp}& 3-layers & Sup. & 3M & 3M & 512 & \href{https://github.com/drorlab/gvp-pytorch}{GitHub: GVP} \\
  & \textsc{Foldseek} \cite{van2024foldseek} & 3Di & Pair. & \textbf{--} & \textbf{--} & \textbf{--} & \href{https://anaconda.org/bioconda/foldseek}{Conda: Foldseek} \\
  & \textsc{TM-align} \cite{zhang2005tmalign} & mean & Pair. & \textbf{--} & \textbf{--} & \textbf{--} & \href{https://anaconda.org/bioconda/tmalign}{Conda: TM-align} \\ 
\bottomrule
\end{tabular}
}
\end{table}

\subsection{Deep Learning Models}
\paragraph{Sequence-Only.}
Sequence-only baselines include both encoder-only and encoder–decoder architectures. Encoder-only models such as \textsc{ESM2} (t30, t33, t36) \cite{lin2023esm2}, \textsc{ESM-1b} \cite{rives2021esm1b}, and \textsc{ProtBert} \cite{elnaggar2021prottrans} are pretrained protein language models using masked language modeling on large sequence corpora. \textsc{Ankh} \cite{elnaggar2023ankh} and \textsc{ProtT5} \cite{elnaggar2021prottrans}, in contrast, adopt encoder–decoder architectures, enabling bidirectional contextualization and autoregressive decoding. While \textsc{TM-vec} \cite{hamamsy2024tmvec} and \textsc{ProstT5} \cite{heinzinger2023prostt5} only require sequence inputs, both incorporate structural inductive signals during training: \textsc{TM-vec} is trained to regress TM-scores, and \textsc{ProstT5} is fine-tuned to translate Foldseek-derived structural tokens.

\paragraph{Sequence–Structure.}
Sequence–structure models combine sequence and structural information in diverse ways. \textsc{SaProt} \cite{su2023saprot} fuses amino acid tokens with Foldseek-derived structural tokens and is trained using multi-modal masked language modeling. \textsc{ProtSSN} \cite{tan2025protssn} integrates \textsc{ESM2} \cite{lin2023esm2} embeddings with geometric graph neural networks, enabling joint sequence–structure representation learning. Both \textsc{ESM-IF1} \cite{hsu2022esm-if1} and \textsc{MIF-ST} \cite{mifst} are inverse folding models: \textsc{ESM-IF1} is pretrained on large-scale backbone recovery, while \textsc{MIF-ST} uses structure-conditioned geometric networks initialized from large protein transformers.

\paragraph{Structure-Only.}
Structure-only baselines rely purely on 3D geometric inputs. \textsc{GVP-GNN} \cite{gvp} is a non-pretrained geometric deep learning model that uses residue type and atomic coordinate features for message passing. 

\subsection{Alignment-based Methods}\label{app:alignment_sec}
\paragraph{Foldseek.}
We employ \textsc{Foldseek}~\cite{van2024foldseek} to evaluate structural similarity between query and target proteins under two alignment modes: \textit{3Di-only} (\texttt{--alignment-type 0}) and \textit{3Di+AA} (\texttt{--alignment-type 2}). To maximize sensitivity, we activate exhaustive pairwise comparison via \texttt{--exhaustive-search}, set a high \texttt{-e} threshold of 1,000, and use \texttt{--min-seq-id 0.0} to allow all sequence identity levels. We retain up to 100,000 alignments per query using \texttt{--max-seqs 100000}, and parallelize computation across available CPU threads. The output alignment scores are used to compute similarity for unsupervised pair-level evaluation (e.g., AUC). \textsc{Foldseek} achieves state-of-the-art tradeoffs between alignment speed and accuracy for large-scale protein structure comparison.

\paragraph{BLASTP.}
We adopt \textsc{BLAST}~\cite{altschul1990blast} as a classical sequence-based alignment baseline. Our setup disables low-complexity masking (\texttt{-seg no}) to preserve short functional regions and sets a permissive $E$-value threshold (\texttt{-evalue 1000000}) to retain weak similarities. Word size is reduced to 2 (\texttt{-word\_size 2}) to improve alignment sensitivity, and a large hit buffer (\texttt{-max\_target\_seqs 100000000}) ensures comprehensive coverage. The output is recorded in tabular format (\texttt{-outfmt 6}), including sequence IDs, identity, alignment length, mismatches, gaps, and bit scores. \textsc{BLAST} is used as a baseline for fragment-level and full-sequence pair similarity evaluation.
\paragraph{TM-align.}
To benchmark structure-only alignment, we apply \textsc{TM-align}~\cite{zhang2005tmalign} on PDB-format query and reference proteins via the default command-line interface (\texttt{TMalign query.pdb ref.pdb}). \textsc{TM-align} returns key statistics, including RMSD, sequence identity, aligned length, and two TM-scores (normalized by query and reference, respectively). We record the average TM-score between the two directions as the final similarity metric for evaluation. \textsc{TM-align} is widely regarded as a reliable tool for structure-based homology assessment, though its quadratic computational complexity limits scalability on large benchmarks.

\section{Detailed Experimental Results}\label{app:add_exp_sec}

Here, we provide detailed experimental results for two tasks: Residue-Level Binary Classification and Fragment-Level Multi-Class Classification. We report all evaluation metrics recorded during the experiments to offer a comprehensive assessment of model performance across different aspects. Please refer to Tables~\ref{app:tab:token_cls_detail_results_1}--\ref{app:tab:fragment_cls_main_results} for the complete results.

\begin{table*}[h]
\caption{Detailed residue-level classification performance across \textbf{BindB} and \textbf{Epi} datasets and data splits. “MP50”, “MP70”, and “MP90” refer to mixed-family splits with 50\%, 70\%, and 90\% sequence identity filtering applied at the full-sequence level. Metrics reported include AUPR, Precision, Recall, F1 scores for negative and positive classes, and Macro-F1.}
\label{app:tab:token_cls_detail_results_1}
\begin{center}
\begin{small}
\resizebox{0.9\textwidth}{!}{
\begin{tabular}{@{}lc|cccccc@{}}
\toprule
\multirow{2}{*}{Metric} & \multirow{2}{*}{Model} & \multicolumn{3}{c}{\textbf{BindB}} & \multicolumn{3}{c}{\textbf{Epi}} \\
\cmidrule(lr){3-5} \cmidrule(lr){6-8}
& & MP50 & MP70 & MP90 & MP50 & MP70 & MP90 \\
\midrule
\multirow{4}{*}{AUPR} & ESM2 t30 & 0.408 & 0.465 & 0.496 & 0.186 & 0.184 & 0.277 \\
& ESM2 t33 & 0.446 & 0.494 & 0.535 & 0.174 & 0.200 & 0.290 \\
& ProtBert & 0.340 & 0.410 & 0.466 & 0.169 & 0.177 & 0.266 \\
& Ankh & 0.421 & 0.487 & 0.527 & 0.167 & 0.215 & 0.270 \\
\midrule
\multirow{4}{*}{precision} & ESM2 t30 & 0.598 & 0.637 & 0.674 & 1.0 & 0.384 & 0.545 \\
& ESM2 t33 & 0.605 & 0.646 & 0.675 & 0.0 & 1.0 & 0.512 \\
& ProtBert & 0.547 & 0.619 & 0.706 & 1.0 & 0.432 & 0.534 \\
& Ankh & 0.634 & 0.660 & 0.677 & 0.0 & 1.0 & 0.571 \\
\midrule
\multirow{4}{*}{Recall} & ESM2 t30 & 0.289 & 0.317 & 0.316 & 0.001 & 0.043 & 0.091 \\
& ESM2 t33 & 0.329 & 0.356 & 0.386 & 0.0 & 0.003 & 0.139 \\
& ProtBert & 0.238 & 0.264 & 0.257 & 0.001 & 0.005 & 0.072 \\
& Ankh & 0.260 & 0.335 & 0.357 & 0.0 & 0.008 & 0.054 \\
\midrule
\multirow{4}{*}{F1-Negative} & ESM2 t30 & 0.987 & 0.986 & 0.986 & 0.958 & 0.960 & 0.968 \\
& ESM2 t33 & 0.987 & 0.987 & 0.987 & 0.958 & 0.961 & 0.968 \\
& ProtBert & 0.986 & 0.986 & 0.986 & 0.958 & 0.961 & 0.968 \\
& Ankh & 0.987 & 0.987 & 0.987 & 0.958 & 0.961 & 0.968 \\
\midrule
\multirow{4}{*}{F1-Positive} & ESM2 t30 & 0.390 & 0.423 & 0.430 & 0.002 & 0.077 & 0.156 \\
& ESM2 t33 & 0.427 & 0.459 & 0.491 & 0.0 & 0.006 & 0.218 \\
& ProtBert & 0.332 & 0.370 & 0.377 & 0.002 & 0.010 & 0.126 \\
& Ankh & 0.369 & 0.444 & 0.483 & 0.0 & 0.002 & 0.098 \\
\midrule
\multirow{4}{*}{Macro-F1} & ESM2 t30 & 0.689 & 0.705 & 0.708 & 0.480 & 0.518 & 0.562 \\
& ESM2 t33 & 0.707 & 0.723 & 0.739 & 0.479 & 0.484 & 0.593 \\
& ProtBert & 0.659 & 0.678 & 0.682 & 0.480 & 0.486 & 0.547 \\
& Ankh & 0.678 & 0.715 & 0.735 & 0.479 & 0.489 & 0.533 \\
\bottomrule
\end{tabular}
}
\end{small}
\end{center}
\vskip -0.1in
\end{table*}

\begin{table*}[t]
\caption{Detailed residue-level classification performance across 
\textbf{Act}, \textbf{BindI}, and \textbf{Evo} datasets and data splits. “MF50” and “MP50” refer to mixed-family splits with 50\% sequence identity filtering applied at the fragment and full-sequence levels, respectively. Metrics reported include AUPR, Precision, Recall, F1 scores for negative and positive classes, and Macro-F1.}
\label{app:tab:token_cls_detail_results_2}
\begin{center}
\begin{small}
\resizebox{\textwidth}{!}{
\begin{tabular}{@{}lc|ccccccccc@{}}
\toprule
\multirow{2}{*}{Metric} & \multirow{2}{*}{Model} & \multicolumn{3}{c}{\textbf{Act}} & \multicolumn{3}{c}{\textbf{BindI}} & \multicolumn{3}{c}{\textbf{Evo}} \\
\cmidrule(lr){3-5} \cmidrule(lr){6-8} \cmidrule(lr){9-11}
& & MF50 & MP50 & Cross & MF50 & MP50 & Cross & MF50 & MP50 & Cross \\
\midrule
\multirow{8}{*}{AUPR} & ESM2 t30 & 0.855 & 0.932 & 0.143 & 0.912 & 0.963 & 0.133 & 0.862 & 0.897 & 0.235 \\
& ESM2 t33 & 0.852 & 0.955 & 0.143 & 0.904 & 0.971 & 0.159 & 0.899 & 0.926 & 0.262 \\
& ProtBert & 0.764 & 0.895 & 0.131 & 0.857 & 0.926 & 0.112 & 0.771 & 0.803 & 0.243 \\
& Ankh base & 0.873 & 0.895 & 0.166 & 0.907 & 0.970 & 0.145 & 0.895 & 0.932 & 0.275 \\
& GVP-GNN & 0.523 & 0.896 & 0.101 & 0.611 & 0.883 & 0.040 & 0.342 & 0.792 & 0.101 \\
& SaProt 35M & 0.688 & 0.905 & 0.114 & 0.807 & 0.927 & 0.230 & 0.724 & 0.772 & 0.272 \\
& SaProt 650M & 0.745 & 0.945 & 0.185 & 0.838 & 0.960 & 0.182 & 0.734 & 0.912 & 0.274 \\
& ProtSSN & 0.465 & 0.917 & 0.156 & 0.801 & 0.907 & 0.095 & 0.715 & 0.895 & 0.227 \\
\midrule

\multirow{8}{*}{Precision} & ESM2 t30 & 0.826 & 0.851 & 0.278 & 0.859 & 0.915 & 0.525 & 0.816 & 0.879 & 0.374 \\
& ESM2 t33 & 0.845 & 0.851 & 0.126 & 0.869 & 0.905 & 0.581 & 0.856 & 0.856 & 0.403 \\
& ProtBert & 0.791 & 0.833 & 0.131 & 0.855 & 0.897 & 0.416 & 0.805 & 0.803 & 0.482 \\
& Ankh base & 0.862 & 0.873 & 0.190 & 0.849 & 0.919 & 0.437 & 0.882 & 0.932 & 0.387 \\
& GVP-GNN & 0.735 & 0.824 & 0.019 & 0.730 & 0.874 & 0.0 & 0.810 & 0.781 & 0.176 \\
& SaProt 35M & 0.818 & 0.879 & 0.132 & 0.813 & 0.902 & 0.634 & 0.819 & 0.841 & 0.382 \\
& SaProt 650M & 0.812 & 0.845 & 0.241 & 0.827 & 0.900 & 0.661 & 0.809 & 0.828 & 0.456 \\
& ProtSSN & 0.523 & 0.835 & 0.241 & 0.818 & 0.887 & 0.379 & 0.790 & 0.815 & 0.452 \\
\midrule

\multirow{8}{*}{Recall}& ESM2 t30     & 0.676 & 0.793 & 0.060 & 0.859 & 0.897 & 0.078 & 0.783 & 0.750 & 0.097 \\
& ESM2 t33     & 0.682 & 0.848 & 0.031 & 0.830 & 0.924 & 0.108 & 0.806 & 0.755 & 0.122 \\
& ProtBert     & 0.565 & 0.750 & 0.020 & 0.694 & 0.839 & 0.048 & 0.610 & 0.597 & 0.009 \\
& Ankh base    & 0.700 & 0.864 & 0.025 & 0.866 & 0.922 & 0.086 & 0.735 & 0.744 & 0.169 \\
& GVP-GNN      & 0.362 & 0.798 & 0.001 & 0.519 & 0.788 & 0.0   & 0.091 & 0.718 & 0.035 \\
& SaProt 35M   & 0.408 & 0.733 & 0.036 & 0.705 & 0.822 & 0.135 & 0.520 & 0.649 & 0.172 \\
& SaProt 650M  & 0.511 & 0.850 & 0.072 & 0.768 & 0.918 & 0.135 & 0.554 & 0.700 & 0.111 \\
& ProtSSN      & 0.209 & 0.801 & 0.014 & 0.705 & 0.788 & 0.029 & 0.507 & 0.852  & 0.034 \\
\midrule

\multirow{8}{*}{F1-Negative} & ESM2 t30     & 0.992 & 0.995 & 0.967 & 0.993 & 0.996 & 0.975 & 0.988 & 0.990 & 0.957 \\
& ESM2 t33     & 0.993 & 0.996 & 0.964 & 0.992 & 0.996 & 0.976 & 0.990 & 0.992 & 0.957 \\
& ProtBert     & 0.991 & 0.994 & 0.967 & 0.989 & 0.994 & 0.974 & 0.984 & 0.986 & 0.960 \\
& Ankh base    & 0.993 & 0.996 & 0.968 & 0.992 & 0.996 & 0.974 & 0.989 & 0.992 & 0.955 \\
& GVP-GNN      & 0.989 & 0.995 & 0.969 & 0.982 & 0.993 & 0.975 & 0.973 & 0.986 & 0.955 \\
& SaProt 35M   & 0.990 & 0.995 & 0.964 & 0.988 & 0.994 & 0.976 & 0.982 & 0.985 & 0.955 \\
& SaProt 650M  & 0.986 & 0.996 & 0.965 & 0.989 & 0.996 & 0.977 & 0.983 & 0.991 & 0.959 \\
& ProtSSN      & 0.988 & 0.995 & 0.969 & 0.988 & 0.993 & 0.975 & 0.982 & 0.991 & 0.960 \\ \midrule

\multirow{8}{*}{F1-Positive} & ESM2 t30     & 0.744 & 0.821 & 0.098 & 0.859 & 0.906 & 0.136 & 0.799 & 0.810 & 0.154 \\
& ESM2 t33     & 0.755 & 0.850 & 0.050 & 0.849 & 0.915 & 0.181 & 0.831 & 0.858 & 0.187 \\
& ProtBert     & 0.659 & 0.789 & 0.035 & 0.766 & 0.867 & 0.086 & 0.694 & 0.701 & 0.017 \\
& Ankh base    & 0.773 & 0.869 & 0.045 & 0.857 & 0.920 & 0.144 & 0.802 & 0.858 & 0.235 \\
& GVP-GNN      & 0.485 & 0.810 & 0.002 & 0.607 & 0.829 & 0.0   & 0.164 & 0.748 & 0.058 \\
& SaProt 35M   & 0.544 & 0.800 & 0.056 & 0.755 & 0.860 & 0.223 & 0.636 & 0.689 & 0.238 \\
& SaProt 650M  & 0.627 & 0.848 & 0.110 & 0.796 & 0.909 & 0.224 & 0.658 & 0.851 & 0.178 \\
& ProtSSN      & 0.329 & 0.818 & 0.026 & 0.757 & 0.839 & 0.053 & 0.618 & 0.833 & 0.062 \\ \midrule

\multirow{8}{*}{Macro-F1} & ESM2 t30     & 0.868 & 0.908 & 0.533 & 0.926 & 0.951 & 0.556 & 0.894 & 0.900 & 0.555 \\
& ESM2 t33     & 0.874 & 0.923 & 0.507 & 0.921 & 0.955 & 0.579 & 0.910 & 0.925 & 0.572 \\
& ProtBert     & 0.825 & 0.892 & 0.501 & 0.878 & 0.931 & 0.530 & 0.839 & 0.843 & 0.489 \\
& Ankh base    & 0.883 & 0.933 & 0.507 & 0.925 & 0.958 & 0.559 & 0.896 & 0.925 & 0.595 \\
& GVP-GNN      & 0.736 & 0.903 & 0.485 & 0.795 & 0.911 & 0.488 & 0.569 & 0.867 & 0.506 \\
& SaProt 35M   & 0.767 & 0.897 & 0.510 & 0.871 & 0.927 & 0.599 & 0.809 & 0.837 & 0.596 \\
& SaProt 650M  & 0.808 & 0.922 & 0.538 & 0.893 & 0.953 & 0.600 & 0.820 & 0.921 & 0.568 \\
& ProtSSN      & 0.658 & 0.906 & 0.498 & 0.873 & 0.911 & 0.514 & 0.800 & 0.912  & 0.511 \\
\bottomrule
\end{tabular}
}
\end{small}
\end{center}
\vskip -0.1in
\end{table*}

\begin{table*}[t]
\caption{Detailed residue-level classification performance across \textbf{Motif} and \textbf{Dom} datasets and data splits. “MF50” and “MP50” refer to mixed-family splits with 50\% sequence identity filtering applied at the fragment and full-sequence levels, respectively. Metrics reported include AUPR, Precision, Recall, F1 scores for negative and positive classes, and Macro-F1.}
\label{app:tab:token_cls_detail_results_3}
\begin{center}
\begin{small}
\resizebox{0.85\textwidth}{!}{
\begin{tabular}{@{}lc|cccccc@{}}
\toprule
\multirow{2}{*}{Metric} & \multirow{2}{*}{Model} & \multicolumn{3}{c}{\textbf{\textbf{Motif}}} & \multicolumn{3}{c}{\textbf{Dom}} \\
\cmidrule(lr){3-5} \cmidrule(lr){6-8}
& & MF50 & MP50 & Cross & MF50 & MP50 & Cross \\

\midrule
\multirow{8}{*}{AUPR} & ESM2 t30 & 0.855 & 0.850 & 0.433 & 0.634 & 0.634 & 0.470\\
& ESM2 t33 & 0.874 & 0.857 & 0.456 & 0.666 & 0.657 & 0.506\\
& ProtBert & 0.779 & 0.796 & 0.348 & 0.591 & 0.592 & 0.508\\
& Ankh base & 0.884 & 0.870 & 0.394 & 0.673 & 0.665 & 0.449\\
& GVP-GNN & 0.661 & 0.736 & 0.329 & 0.560 & 0.557 & 0.468\\
& SaProt 35M & 0.767 & 0.784 & 0.408 & 0.574 & 0.584 & 0.525\\
& SaProt 650M & 0.802 & 0.841 & 0.441 & 0.642 & 0.640 & 0.564\\
& ProtSSN & 0.716 & 0.765 & 0.390 & \textbf{--} & \textbf{--} & \textbf{--}\\
\midrule

\multirow{8}{*}{Precision} & ESM2 t30 & 0.824 & 0.802 & 0.510 & 0.648 & 0.644 & 0.496 \\
& ESM2 t33 & 0.851 & 0.795 & 0.566 & 0.661 & 0.634 & 0.530 \\
& ProtBert & 0.784 & 0.793 & 0.472 & 0.636 & 0.596 & 0.588 \\
& Ankh base & 0.846 & 0.817 & 0.499 & 0.674 & 0.646 & 0.494 \\
& GVP-GNN & 0.748 & 0.756 & 0.329 & 0.591 & 0.557 & 0.519 \\
& SaProt 35M & 0.821 & 0.783 & 0.485 & 0.632 & 0.615 & 0.548 \\
& SaProt 650M & 0.841 & 0.818 & 0.504 & 0.635 & 0.656 & 0.572 \\
& ProtSSN & 0.772 & 0.775 & 0.390 & \textbf{--} & \textbf{--} & \textbf{--} \\
\midrule

\multirow{8}{*}{Recall}& ESM2 t30 & 0.775 & 0.731 & 0.432 & 0.433 & 0.423 & 0.360\\
& ESM2 t33 & 0.748 & 0.861 & 0.384 & 0.467 & 0.478 & 0.367\\
& ProtBert & 0.678 & 0.592 & 0.231 & 0.353 & 0.420 & 0.138\\
& Ankh base & 0.789 & 0.831 & 0.303 & 0.467 & 0.490 & 0.280\\
& GVP-GNN & 0.525 & 0.669 & 0.453 & 0.344 & 0.309 & 0.087\\
& SaProt 35M & 0.582 & 0.954 & 0.411 & 0.322 & 0.840 & 0.349\\ 
& SaProt 650M & 0.615 & 0.960 & 0.350 & 0.472 & 0.414 & 0.444\\ 
& ProtSSN & 0.550 & 0.676 & 0.365 & \textbf{--} & \textbf{--} & \textbf{--}\\
\midrule

\multirow{8}{*}{F1-Negative} & ESM2 t30 & 0.972 & 0.961 & 0.946 & 0.839 & 0.849 & 0.738\\
& ESM2 t33 & 0.972 & 0.962 & 0.951 & 0.844 & 0.848 & 0.752\\
& ProtBert & 0.963 & 0.952 & 0.945 & 0.834 & 0.837 & 0.779\\
& Ankh base & 0.974 & 0.963 & 0.946 & 0.847 & 0.853 & 0.748\\
& GVP-GNN & 0.954 & 0.953 & 0.924 & 0.836 & 0.837 & 0.774\\
& SaProt 35M & 0.961 & 0.954 & 0.944 & 0.832 & 0.840 & 0.761\\ 
& SaProt 650M & 0.964 & 0.960 & 0.946 & 0.837 & 0.850 & 0.765\\
& ProtSSN & 0.956 & 0.954 & 0.944 & \textbf{--} & \textbf{--} & \textbf{--}\\
\midrule

\multirow{8}{*}{F1-Positive} & ESM2 t30 & 0.799 & 0.765 & 0.467 & 0.519 & 0.510 & 0.417\\
& ESM2 t33 & 0.796 & 0.774 & 0.457 & 0.547 & 0.545 & 0.433\\
& ProtBert & 0.727 & 0.681 & 0.310 & 0.454 & 0.4936 & 0.223\\
& Ankh base & 0.817 & 0.779 & 0.377 & 0.552 & 0.557 & 0.357\\
& GVP-GNN & 0.618 & 0.710 & 0.399 & 0.435 & 0.408 & 0.149\\
& SaProt 35M & 0.681 & 0.709 & 0.445 & 0.427 & 0.462 & 0.427\\ 
& SaProt 650M & 0.710 & 0.754 & 0.414 & 0.542 & 0.508 & 0.500\\
& ProtSSN & 0.642 & 0.720 & 0.412 & \textbf{--} & \textbf{--} & \textbf{--}\\
\midrule

\multirow{8}{*}{Macro-F1} & ESM2 t30 & 0.885 & 0.863 & 0.707 & 0.679 & 0.680 & 0.578\\
& ESM2 t33 & 0.884 & 0.868 & 0.704 & 0.696 & 0.697 & 0.593\\
& ProtBert & 0.845 & 0.816 & 0.628 & 0.644 & 0.665 & 0.501\\
& Ankh base & 0.895 & 0.871 & 0.662 & 0.700 & 0.745 & 0.552\\
& GVP-GNN & 0.786 & 0.831 & 0.661 & 0.636 & 0.623 & 0.462\\
& SaProt 35M & 0.821 & 0.832 & 0.695 & 0.629 & 0.651 & 0.594\\
& SaProt 650M & 0.837 & 0.857 & 0.680 & 0.689 & 0.679 & 0.632\\
& ProtSSN & 0.799 & 0.837 & 0.678 & \textbf{--} & \textbf{--} & \textbf{--}\\

\bottomrule
\end{tabular}
}
\end{small}
\end{center}
\vskip -0.1in
\end{table*}

\begin{table*}[t]
\caption{Detailed fragment-level classification results on “MF50” split across \textbf{Act}, \textbf{BindI}, \textbf{Evo}, and \textbf{Motif} datasets. “MF50” refers to mixed-family splits with 50\% sequence identity filtering applied at the fragment level. Metrics reported include Accuracy, Precision, Recall, Macro-F1, and Matthews Correlation Coefficient (MCC).}
\label{app:tab:fragment_cls_main_results}
\begin{center}
\begin{small}
\resizebox{0.6\textwidth}{!}{
    \begin{tabular}{@{}lc|cccc@{}}
    \toprule
    Metric & Model & \textbf{Act} & \textbf{BindI} & \textbf{Evo} & \textbf{Motif} \\
    \midrule
    \multirow{8}{*}{Accuracy} & ESM2 t30 & 0.819 & 0.937 & 0.853 & 0.884 \\  
    & ESM2 t33 & 0.814 & 0.934 & 0.841 & 0.906 \\
    & ProtBert & 0.736 & 0.927 & 0.828 & 0.884 \\
    & Ankh base & 0.824 & 0.920 & 0.866 & 0.901 \\
    & GVP-GNN & 0.907 & 0.972 & 0.914 & 0.807 \\
    & SaProt 35M & 0.928 & 0.976 & 0.939 & 0.901 \\
    & SaProt 650M & 0.928 & 0.986 & 0.950 & 0.927 \\
    & ProtSSN & 0.891 & 0.972 & 0.915 & 0.914 \\
    \midrule
     
    \multirow{8}{*}{Precision} & ESM2 t30 & 0.659 & 0.834 & 0.681 & 0.458 \\  
    & ESM2 t33 & 0.603 & 0.755 & 0.682 & 0.547 \\
    & ProtBert & 0.618 & 0.838 & 0.644 & 0.455 \\
    & Ankh base & 0.661 & 0.733 & 0.727 & 0.508 \\
    & GVP-GNN & 0.826 & 0.901 & 0.763 & 0.387 \\
    & SaProt 35M & 0.810 & 0.943 & 0.857 & 0.509 \\
    & SaProt 650M & 0.830 & 0.968 & 0.868 & 0.546 \\
    & ProtSSN & 0.773 & 0.940 & 0.804 & 0.564 \\
    \midrule
    
    \multirow{8}{*}{Recall} & ESM2 t30 & 0.670 & 0.819 & 0.684 & 0.461 \\  
    & ESM2 t33 & 0.634 & 0.775 & 0.682 & 0.543 \\
    & ProtBert & 0.636 & 0.794 & 0.646 & 0.458 \\
    & Ankh base & 0.665 & 0.732 & 0.729 & 0.501 \\
    & GVP-GNN & 0.833 & 0.882 & 0.768 & 0.371 \\
    & SaProt 35M & 0.823 & 0.929 & 0.858 & 0.505 \\
    & SaProt 650M & 0.830 & 0.956 & 0.875 & 0.562 \\
    & ProtSSN & 0.774 & 0.948 & 0.807 & 0.556 \\
    \midrule
    
    \multirow{8}{*}{Macro-F1} & ESM2 t30 & 0.647 & 0.809 & 0.667 & 0.457 \\  
    & ESM2 t33 & 0.605 & 0.753 & 0.669 & 0.542 \\
    & ProtBert & 0.609 & 0.790 & 0.627 & 0.452 \\
    & Ankh base & 0.647 & 0.718 & 0.716 & 0.499 \\
    & GVP-GNN & 0.822 & 0.884 & 0.757 & 0.370 \\
    & SaProt 35M & 0.807 & 0.931 & 0.849 & 0.504 \\
    & SaProt 650M & 0.825 & 0.957 & 0.863 & 0.552 \\
    & ProtSSN & 0.764 & 0.931 & 0.793 & 0.556 \\
    \midrule
    
    \multirow{8}{*}{MCC} & ESM2 t30 & 0.815 & 0.926 & 0.852 & 0.875 \\  
    & ESM2 t33 & 0.810 & 0.922 & 0.840 & 0.898 \\
    & ProtBert & 0.731 & 0.914 & 0.827 & 0.875 \\
    & Ankh base & 0.821 & 0.906 & 0.865 & 0.892 \\
    & GVP-GNN & 0.906 & 0.967 & 0.913 & 0.791 \\
    & SaProt 35M & 0.926 & 0.971 & 0.938 & 0.892 \\
    & SaProt 650M & 0.926 & 0.984 & 0.950 & 0.921 \\
    & ProtSSN & 0.889 & 0.967 & 0.915 & 0.907 \\
    \bottomrule
    
    \end{tabular}
    }
\end{small}
\end{center}
\vskip -0.1in
\end{table*}

\end{document}